\documentclass[runningheads]{llncs}
\usepackage{graphicx}
\usepackage{amsmath,amssymb} 
\usepackage{color}
\usepackage{vwcol}  
\usepackage[linesnumbered,lined,boxed,commentsnumbered]{algorithm2e}
\newcommand{\nosemic}{\SetEndCharOfAlgoLine{\relax}}

\usepackage[width=122mm,left=12mm,paperwidth=146mm,height=193mm,top=12mm,paperheight=217mm]{geometry}

\begin{document}
\pagestyle{headings}
\mainmatter

\title{Massively Parallel Video Networks}

\titlerunning{Massively Parallel Video Networks}

\authorrunning{Carreira, P{\u a}tr{\u a}ucean et al.}

\author{Jo{\~ a}o Carreira$^{\dagger,1}$, Viorica P{\u a}tr{\u a}ucean$^{\dagger,1}$, Laurent Mazare$^1$, \\ Andrew Zisserman$^{1,2}$, Simon Osindero$^1$}

\institute{
$^1$DeepMind \\ $^2$Department of Engineering Science, University of Oxford\\
\email{\{joaoluis, viorica, mazare, zisserman, osindero\}@google.com} \\
$^{\dagger}$shared first author}

\newcommand{\informationlatency}{information latency}
\newcommand{\predictionlag}{prediction latency}

\maketitle

\begin{abstract}
We introduce a class of causal video understanding models that aims to improve efficiency of video processing by maximising throughput, minimising latency, and reducing the number of clock cycles. 
Leveraging operation pipelining and multi-rate clocks, these models perform a minimal amount of computation (e.g. as few as four convolutional layers) for each frame per timestep to produce an output. 
The models are still very deep, with dozens of such operations being performed but in a pipelined fashion that enables depth-parallel computation. We illustrate the proposed principles by applying them to existing image architectures and analyse their behaviour on two video tasks: action recognition and human keypoint localisation. The results show that a significant degree of parallelism, and implicitly speedup, can be achieved with little loss in performance.

\end{abstract}

\section{Introduction}

There is a rich structure in videos that is neglected when treating them as a set of still images. Perhaps the most explored benefit of videos is the ability to improve performance by aggregating information over multiple frames~\cite{jampani17vpn,pfister2015flowing,zhuflowguided}, which enforces temporal smoothness and reduces the uncertainty in tasks that are temporal by nature, e.g., change detection~\cite{Stent-RSS-16}, computing optical flow~\cite{IMKDB17}, resolving action ambiguities (standing up/sitting down)~\cite{carreira2017cvpr} etc. An underexplored direction, however, is the ability to improve the processing efficiency. In this paper, we focus on this aspect in the context of the causal, frame-by-frame operation mode that is relevant for real-time applications, and show how to transform slow models to ones that can run at frame rate with negligible loss of accuracy. 

Most existing state-of-the-art computer vision systems, such as object detectors~\cite{NIPS2015_5638,DBLP:conf/cvpr/RedmonF17,he2017mask}, process video frames independently: each new frame goes through up to one hundred convolutional layers before the output is known and another frame can be processed. This sequential operation in both depth and time can pose several problems: it can limit the rate at which predictions can be made, it can increase the minimum latency with which good predictions are available, and it can also lead to under-utilisation of hardware resources.

General-purpose computer processors encounter the same challenge when executing sequences of program instructions and address it with efficient pipelining strategies, that enable parallel computations. 
This also resembles the operation mode of biological neurons, which are not tremendously fast, but come in large numbers and operate in a massively parallel fashion~\cite{Zeki20140174}. 
Our proposed design employs similar pipelining strategies, and we make four contributions:
first, we propose pipelining schemes tailored to sequence models (we call this \emph{predictive depth-parallelism}); 
second, we show how such architectures can be augmented using \emph{multi-rate clocks} and how they benefit from skip connections.
These designs can be incorporated into any deep image architecture, to increase their throughput (frame rate) by a large factor (up to 10x in our experiments) when applied on videos. However they may also negatively impact accuracy. To reduce this impact, and as a third contribution, we show that it is  possible to get better parallel models by {\em distilling} them from sequential ones and, as a final contribution, we explore other wiring patterns -- {\em  temporal filters and feedback} -- that improve the expressivity of the resulting models. Collectively, this results in video networks with the ability to make accurate predictions at very high frame rates.

We will discuss related work in the next section. Then, we will move on to describe predictive depth-parallelism, multi-rate clocks and our other technical contributions in sec. \ref{sec:model}. In sec. \ref{sec:experiments} we present our main experiments on two types of prediction tasks with different latency requirements: human keypoint localisation (which requires predicting a dense heatmap for each frame in a video); and action recognition (where a single label is predicted for an entire video clip), before the paper concludes.

\section{Related work}
\label{sec:related_work}

The majority of existing video models rely on image models~\cite{inception,huang2017densely,Simonyan14c} executed frame-by-frame, the main challenge being to speed up the image models to process sequentially 25 frames per second. This can be achieved by simplifying the models, either by identifying accurate architectures with fewer parameters \cite{howard2017mobilenets}, by pruning them  post-training~\cite{Chen:2015:CNN:3045118.3045361}, or by using low-bit representation formats~\cite{courbariaux+al-2016-binarized}. All of these can be combined with our approach. 

A different type of model incorporates recurrent connections~\cite{Srivastava:2015:ULV:3045118.3045209,PatrauceanHC16,Tokmakov_2017_ICCV} for propagating information between time steps~\cite{PatrauceanHC16,Tokmakov_2017_ICCV}. One simple propagation scheme, used by Zhu et al \cite{zhu2017deep} proposed periodically warping old activations given fresh external optical flow as input, rather than recomputing them. Our pipelining strategy has the advantage that it does not require external inputs nor special warping modules. Instead, it places the burden on learning. 

There are also models that consider the video as a volume by stacking the frames and applying 3D convolutions to extract spatio-temporal features~\cite{Tran:2015:LSF:2919332.2919929,carreira2017cvpr}. These models scale well and can be trained on large-scale datasets~\cite{caba2015activitynet,gu2017,DBLP:journals/corr/KayCSZHVVGBNSZ17} due to the use of larger temporal convolution strides at deeper layers. Although they achieve state-of-the-art performance on tasks such as action recognition, these methods still use purely sequential processing in depth (all layers must execute before proceeding to a next input). Moreover, they are not causal -- the 3D convolutional kernels extract features from future frames, which makes it challenging to use these models in real-time.

In the causal category, a number of hierarchical architectures have been proposed around the notion of \emph{clocks}, attaching to each module a possibly different clock rate, yielding temporally multi-scale models that scale better to long sequences~\cite{pmlr-v32-koutnik14}. 
The clock rates can be hard-coded~\cite{DBLP:conf/icml/VezhnevetsOSHJS17} or learnt from  data~\cite{neil2016phased}. Some recent models ~\cite{Shelhamer2016,figurnov2017cvpr} activate different modules of the network based on the temporal and spatial variance of the inputs, respectively, yielding adaptive clocks. There is also a group of time-budget methods that focuses on reducing latency. If the available time runs out before the data has traversed the entire network, then emergency exits are used to output whatever prediction have been computed thus far~\cite{karayev14cvpr,Mathe_2016_CVPR}. This differs from our approach which aims for constant low-latency output.

Ideas related to pipelining were discussed in \cite{Shelhamer2016}; a recent paper also proposed pipelining strategies for speeding up backpropagation for faster training in distributed systems~\cite{Petrowski:1993:PAP:2325858.2328362,pipelined-back-propagation-for-context-dependent-deep-neural-networks,jaderberg2017decoupled}. Instead, we focus on pipelining at inference time, to reduce latency and maximise frame rate. 

\section{Efficient online video models}
\label{sec:model}

\begin{figure}[t!]
\caption{Illustration of a standard sequential video model that processes frames independently, and depth-parallel versions. The horizontal direction represents the time and the vertical direction represents the depth of the network. The throughput of the basic image model depicted in \textbf{(a)} can be increased for real-time video processing using depth-parallelisation, shown in \textbf{(b)}. This makes it possible to, given a new frame, process all layers in parallel, increasing throughput if parallel resources are available. But this also introduces a delay of a few frames -- in this example, the output at time $t$ corresponds to the input at time $t-3$. It is possible to train the network to anticipate the correct output in order to reduce the latency \textbf{(c)}. This task can be made easier if the model has skip-connections, as illustrated in \textbf{(d)} -- this way the model has access to some fresh features (albeit these fresh features have limited computational depth).}
\centering
\includegraphics[width=0.85\textwidth]{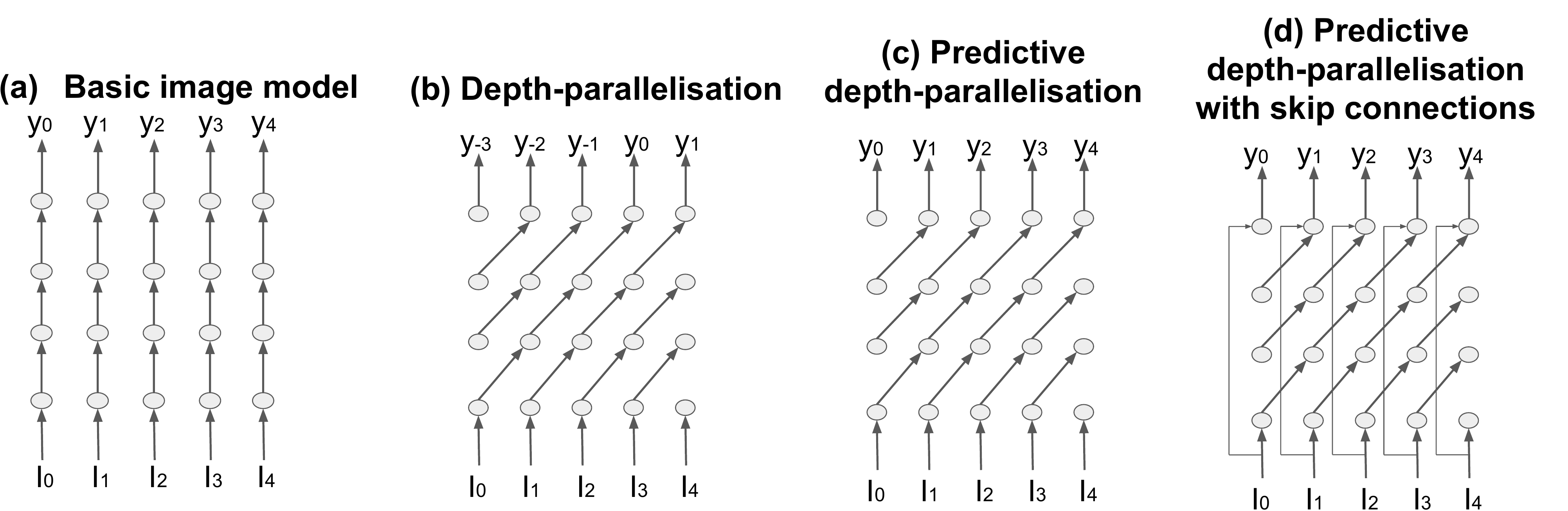}
\label{fig:architecture-lag}
\end{figure}

Consider the directed graph obtained by unrolling a video model with $n$ layers over time (see fig.~\ref{fig:architecture-lag}), where the layers of the network are represented by the nodes and the activations transferred between layers are represented by the edges of the graph. All the parameters are shared across time steps. Edges create dependencies in the computational graph and require sequential processing. Video processing can be efficiently parallelised in the offline case, by processing different frames in different computing cores, but not in the online case.

\noindent\textbf{Depth-parallel networks.} In basic depth-sequential video models, the input to each layer is the output of the previous layer at the same time step, and the network outputs a prediction only after all the layers have processed in sequence the current frame; see fig.~\ref{fig:architecture-lag} (a). In the proposed design,  every layer in the network processes its input, passes the activations to the next layer, and immediately starts processing the next input available, without waiting for the whole network to finish computation for the current frame; fig.~\ref{fig:architecture-lag} (b). This is achieved by substituting in the unrolled graph the vertical edges by diagonal ones, so the input to each layer is still the output from the previous layer, as usual, but \textit{from the previous time step}. This makes it possible to process all layers at one time step in parallel, given enough computing cores, since there are no dependencies between them.

\noindent\textbf{Latency and throughput}. We define \emph{computational latency}, or just latency, as the time delay between the moment when a frame is fed to the network and the moment when the network outputs a prediction for that frame. It is the sum of the execution times of all layers for processing a frame. We consider \emph{throughput} as the output rate of a network, i.e. for how many frames does the network output predictions for in a time unit. For the sequential model, throughput is roughly the inverse of the computational latency, hence the deeper the model, the higher the computational latency and the lower the throughput. Here resides a quality of the proposed depth-parallel models: irrespective of the depth, the model can now make predictions at the rate of its slowest layer.

It is useful to also consider the concepts of \emph{\informationlatency} \ as the number of frames it takes before the input signal  reaches the output layer along the network's shortest path. For example, in fig.~\ref{fig:architecture-lag}, the \informationlatency \ for the video model illustrated in (a) is 0, and for the model in (b) it is equal to 3. We define \emph{\predictionlag} \ as the displacement measured in frames between the moment when a network receives a frame and the moment when the network tries to emit the corresponding output. The \predictionlag \ is a training choice and can have any value. Whenever the \predictionlag \ is smaller than the \informationlatency, the network must make a prediction for an input that it did not process yet completely.

For most of our experiments with depth-parallel models we used a \predictionlag\ of zero
based on the assumption that videos may be predictable over short horizons and we train the network to compensate for the delay in its inputs and operate in a predictive fashion; see fig.~\ref{fig:architecture-lag} (c). But the higher the \informationlatency, the more challenging it is to operate with \predictionlag\ of zero. We employ temporal skip connections to minimise the \informationlatency \ of the different layers in the network, as illustrated in 
 fig.~\ref{fig:architecture-lag} (d). This provides fresher (but shallower) inputs to deeper layers. We term this overall paradigm \emph{predictive depth-parallelism}. 
We experimented thoroughly with the setting where \predictionlag\ is zero and also report results with slightly higher values (e.g. 2 frames).

\begin{figure}[t!]
\begin{center}
\includegraphics[width=0.6\linewidth]{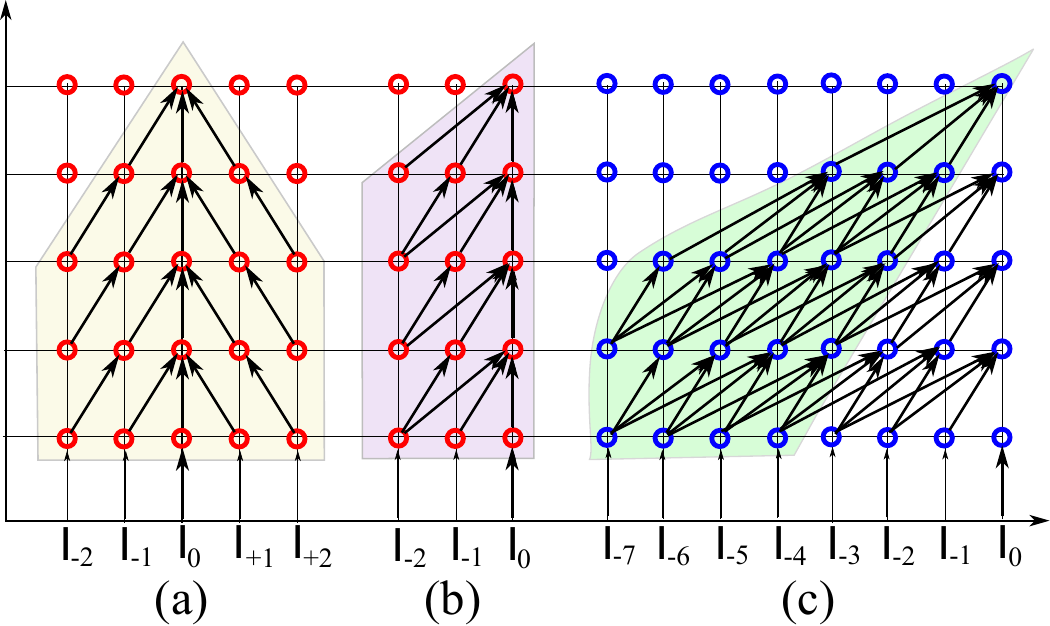}
\end{center}
\caption{Temporal receptive fields of: \textbf{(a)} standard; \textbf{(b)} causal; and \textbf{(c)} pipelined models.
\label{fig:latency}}
\end{figure}

\noindent \textbf{Pipelined operations and temporal receptive field.} Depth-parallelism  has implications regarding the temporal receptive field of the network. 
In any standard neural network, by design, the temporal receptive field of a layer, i.e. the frames its input data comes from, is always a subset of the temporal receptive field of the next deeper layer in the network, resulting in a symmetric triangular shape; see fig.~\ref{fig:latency} (a). 
Stacked temporal convolutions and pooling layers are used for increasing the temporal visual field for deeper layers. In causal models the temporal receptive field is a right-angled triangle -- no layer in the network has access to future frames; see fig.~\ref{fig:latency} (b). In the proposed design, the temporal receptive field along the depth of the network has a skewed triangular shape, the shallower layers having access to frames that the deeper layers cannot yet see (\informationlatency). For example in fig.~\ref{fig:latency} (c), the latest frame that the deepest layer can see at time $t=0$ is the frame $\text{I}_{-4}$, assuming a temporal kernel of 3, which, since we define a \predictionlag \ of zero, means it must predict the output 4 frames in advance.  Adding temporal skip connections reduces the \informationlatency; at the extreme the receptive field becomes similar to the causal one, bringing it to zero.

\begin{figure}[h]
\caption{\textbf{Left}: neural networks with three parallel subnetworks of two layers and two parallel subnetworks of three layers. \textbf{Right}: sequential-to-parallel distillation, the additional loss $L(\hat a, a)$ leverages intermediate activations of the pre-trained sequential model. \label{fig:sub-distillation}}
\centering
\includegraphics[width=0.8\textwidth]{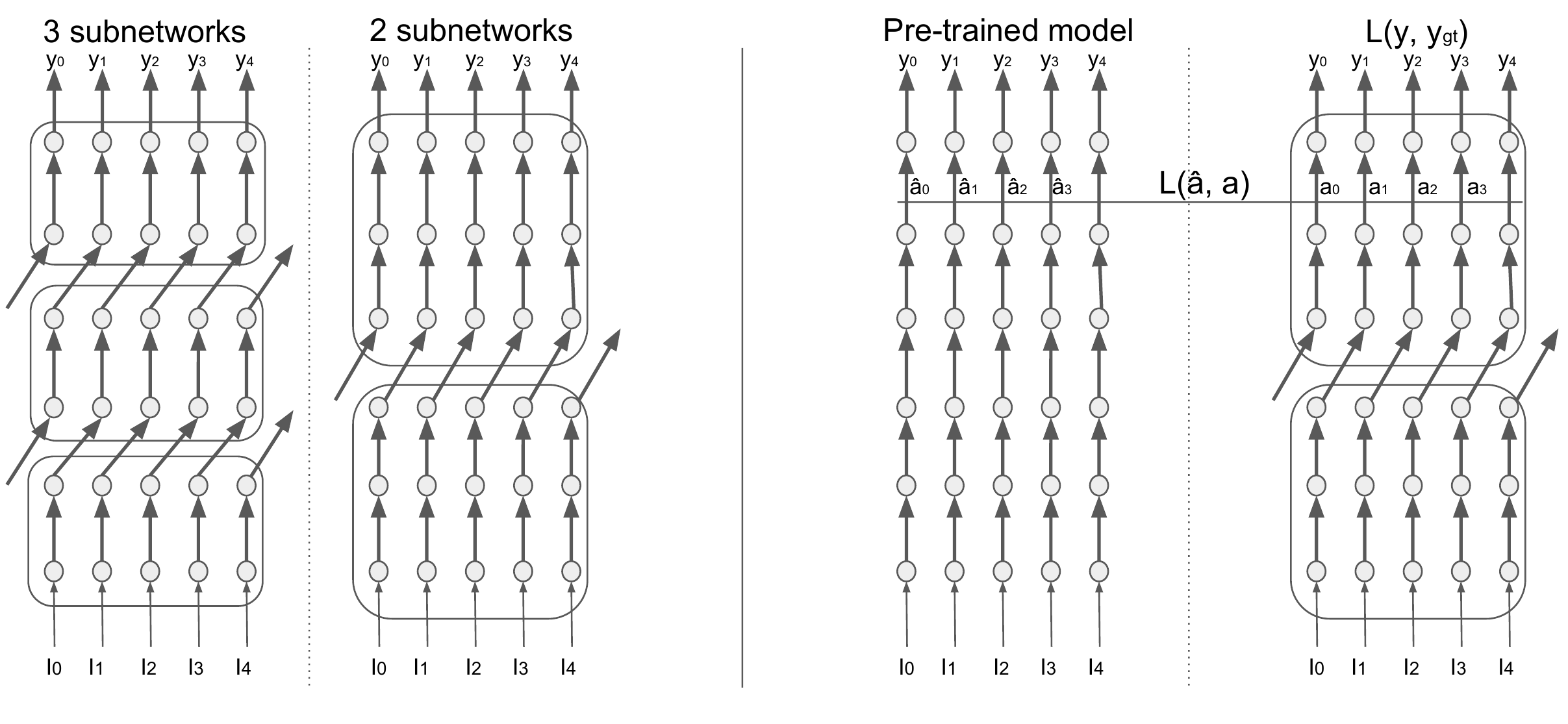}
\end{figure}

\noindent \textbf{Levels of parallelism.} 
For simplicity, the proposed design ideas were illustrated in fig.~\ref{fig:architecture-lag} using the ``extreme'' models, i.e.: (a) which is fully-sequential (with only vertical edges); and (b-c): which are fully parallel  (lacking any vertical edge). However, there is a whole space of semi-parallel models in between, which makes it possible to trade off accuracy and efficiency.
A simple strategy to transform an image model with a linear-chain layer-architecture into a semi-parallel video model is to traverse the network starting from the first layer, and group together contiguous layers into sequential blocks of $k$ layers that we will call \emph{parallel subnetworks} and which can execute independently -- see the two diagrams on the right side of fig.~\ref{fig:sub-distillation}, left; basic pseudocode is given in the supp. material.

\subsection{Multi-rate clocks}
Features extracted deeper in a neural network tend to be more abstract and to vary less over time  \cite{Shelhamer2016}, obeying the so-called \emph{slowness principle}~\cite{Wiskott:2002:SFA:638940.638941} -- fast varying observations can be explained by slow varying latent factors. For example, when tracking a non-rigid moving object, the contours, which are shallow features, change rapidly, but the identity of the object typically does not change at all.  Since not all features change at the same rate as the input rate, it is then possible to reduce computation by reusing, and not recomputing, the deeper, more abstract, features. This can be implemented by having multi-rate clocks: whenever the clock of a layer does not tick, that layer does not compute activations, instead it reuses the existing ones.  
3D ConvNets implement this principle by using temporal strides but does not keep state and hence cannot efficiently operate frame-by-frame. In our recurrent setting, multi-rate clocks can be implemented by removing nodes from the unrolled graph and preserving an internal state to cache outputs until the next slower-ticking layer can consume them. We used a set of fixed rates in our models, typically reducing clock rates by a factor of two whenever spatial resolution is halved. Instead of just using identity to create the internal state as we did, one could use any spatial recurrent module (conv. versions of vanilla RNNs or LSTMs). This design is shown in fig.~\ref{fig:wiring-patterns} (d).

For pixelwise prediction tasks, the state tensors from the last layer of a given spatial resolution are also passed through skip connections, bilinearly upsampled and concatenated as input to the dense prediction head, similar to the skip connections in FCN models~\cite{Shelhamer:2017:FCN:3069214.3069246}, but arise from previous time steps \footnote{More sophisticated trainable decoders, such as those in U-Nets~\cite{Ronneberger2015}, could also be used in a similar pipelined fashion as the encoder.}.

\subsection{Temporal filters and feedback}
\begin{figure}[h!]
\caption{Basic image models (left) can be extended along the temporal domain using different patterns of connectivity. 
Temporalisation adds additional inputs to the different computation nodes, increasing their temporal receptive field. Feedback re-injects past high-level activations to the bottom of the network. Both connectivity patterns aim to improve the expressivity of the models. For increasing throughput, having multi-rate clocks avoids always computing deeper activations (here shown for a temporal model), and instead past activations are copied periodically.}
\label{fig:wiring-patterns}
\centering
\includegraphics[width=1.0\textwidth]{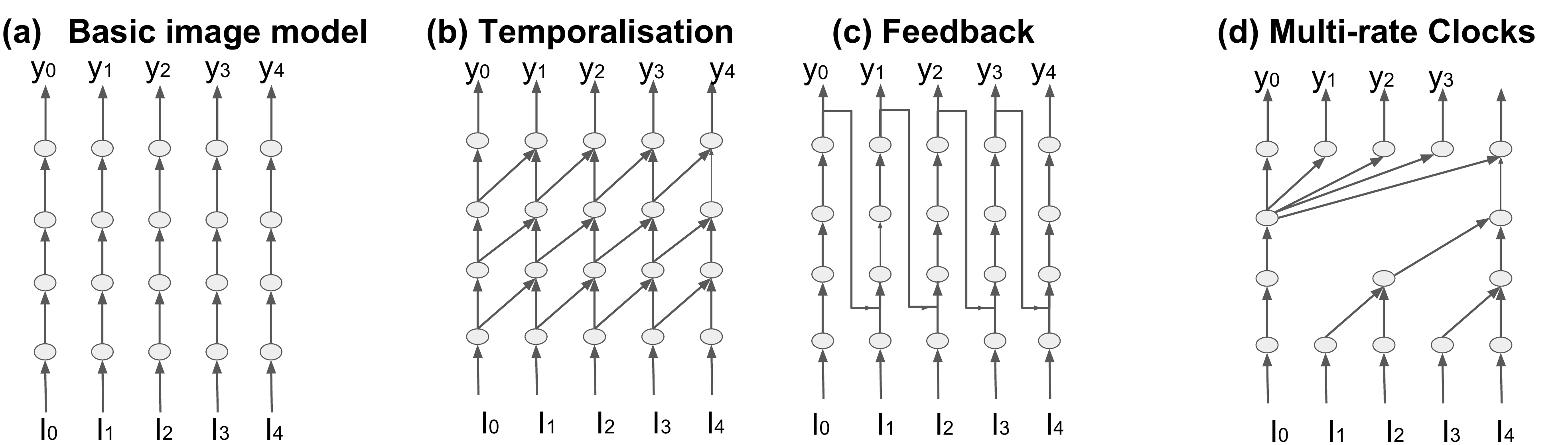}
\end{figure}

The success of depth-parallelism and multi-rate clocks depends on the network being able to learn to compensate for otherwise delayed, possibly stale inputs, which may be feasible since videos are quite redundant and scene dynamics are predictable over short temporal horizons. One way to make learning easier would seem to be by using units with temporal filters. These have shown their worth in a variety of video models~\cite{simonyan2014two,Tran:2015:LSF:2919332.2919929,carreira2017cvpr}. We illustrate the use of temporal filters in fig.~\ref{fig:wiring-patterns}, (b) as \textit{temporalisation}. Interestingly, depth-paralellisation by itself also induces temporalisation in models with skip connections.

For dense predictions tasks, we experimented with adding a feedback connection -- the outputs of the previous frame are fed as inputs to the early layers of the network (e.g. stacking them with the output of the first conv. layer). The idea is that previous outputs provide a simple starting solution with rich semantics which can be refined in few layers -- similar to several recent papers~\cite{carreira2016human,belagiannis2017recurrent,li2016iterative,stollenga2014deep,zamir2017feedback}. This design is shown in fig.~\ref{fig:wiring-patterns}, (c).

\subsection{Sequential-to-parallel ``distillation''}
The proposed parallel models reduce latency, but their computational depth for the current frame at the moment where they produce an output is also reduced compared to their fully sequential counterparts; additionally they are designed to re-use features from previous states through the multi-rate clocks mechanism. These properties typically make learning more difficult. 
In order to improve the accuracy of our parallel models, we adopt a strategy similar to distillation ~\cite{hinton2015distilling}, or to Ladder networks~\cite{Rasmus:2015:SLL:2969442.2969635}, wherein a \emph{teacher} network is privileged relative to a \emph{student} network, either due to having a greater capacity or (in the case of Ladder networks) access to greater amounts of information.

In our case, we consider the sequential model as the teacher, since all of its layers always have access to fresh features extracted from the current frame. We first train a causal fully-sequential model with the same overall architecture as the parallel model. Then we modify the loss of the parallel model to encourage its activations to match those of the sequential model for some given layers, while still minimising the original classification error, such that it predicts how the abstract features would have looked, had the information from the current frame been available. This is illustrated for one layer on the right side of fig.~\ref{fig:sub-distillation}. In our experiment we used the average of this new loss over $m=3$ layers.
The overall loss $L_d$ with distillation is:
\begin{equation*}
L_d = L(y, y_{gt}) + \lambda \sum_{i=1}^m \frac{1}{n_i} \left\lVert \hat a^{(i)} - a^{(i)} \right\rVert^2
\end{equation*}
where $L(y, y_{gt})$ is the initial cross-entropy loss between the predictions of the parallel network $y$ and the ground truth $y_{gt}$, and the second term is the normalised Euclidean distance between the activations of the pre-trained sequential model $\hat a^{(i)}$ for layer $i$ and the activation of the parallel model $a^{(i)}$ for the same layer; $n_i$ denotes the number of feature channels of layer $i$. A parameter $\lambda$ is used to weight the two components of the new loss. We set $\lambda=1$ for the dense keypoint prediction and $\lambda=100$ for action recognition.

\section{Experiments}
\label{sec:experiments}

We applied the proposed principles starting from two popular image classification models: a 54 layer DenseNet~\cite{huang2017densely} and Inception~\cite{inception}, which has 22 conv. layers. We chose these models due to their differences in connectivity. Inception has some built-in parallelism due to the parallel branches in the Inception blocks. DenseNet has no parallelism and instead has dense skip connections within blocks, which helps reduce \informationlatency~when parallelised. Full details on the architectures are provided in the supp. material.

We instantiated a number of model variations using the principles set in the previous section. In all cases we are interested in the online, causal setting (i.e. no peeking into the future), where efficiency matters the most. In the majority of the experiments we trained models with 0 \predictionlag~(e.g. the output at time t should correspond to the input at time t), the most challenging setting. We name pipelined DenseNet models as Par-DenseNet and Inception-based models as Par-Inception.

For evaluation, we considered two tasks having different latency and throughput requirements: (1) action classification, where the network must output only one label prediction for the entire video sequence, and (2) human keypoint localisation, where the network must output dense per-frame predictions for the locations of human joints -- in our case spatial heatmaps for the keypoints of interest (see fig.~\ref{fig:kinetics_pose_samples}).

The dataset for training and evaluation in all cases  was miniKinetics \cite{Xie2017RethinkingSF}, which has 80k training videos and 5k test videos. MiniKinetics is a subset of the larger Kinetics \cite{DBLP:journals/corr/KayCSZHVVGBNSZ17}, but more pratical when studying many factors of variation. For heatmap estimation we populated miniKinetics automatically with poses from a state-of-the-art 2D pose estimation method~\cite{45946} -- that we will call \emph{baseline} from now on -- and used those as ground truth. This resulted in a total of 20 million training frames \footnote{This is far higher than the largest 2D pose video dataset, PoseTrack \cite{PoseTrack}, which has just 20k annotated frames, hardly sufficient for training large video models from scratch (although cleanly annotated instead of automatically).}.

\subsection{Action recognition}
\label{sec:experiments-actionrecognition}
For this task we experimented with three levels of depth-parallelism for both architectures: fully sequential, 5, and 10 parallel subnetworks for Par-Inception models and fully sequential, 7, and 14 parallel subnetworks for Par-DenseNet models. Table \ref{tab:sparse_results} presents the results in terms of Top-1 accuracy on miniKinetics. The accuracy of the original I3D model \cite{carreira2017cvpr} on miniKinetics is $78.3\%$, as reported in~\cite{Xie2017RethinkingSF}. This model is non-causal, but otherwise equivalent to the fully sequential version of our Par-Inception \footnote{Note that this was pre-trained using ImageNet, hence it has a significant advantage over all our models that are trained from scratch.}. 

There is a progressive degradation in performance as more depth-parallelism is added, i.e. as the models become faster and faster, illustrating the trade-off between speedup and accuracy. One possible explanation is the narrowing of the temporal receptive field, shown in fig.~\ref{fig:latency}. The activations of the last frames in each training clip do not get to be processed by the last classifier layer, which is equivalent to training on shorter sequences -- a factor known to impact negatively the classification accuracy. We intend to increase the length of the clips in future work to explore this further. Promisingly, the loss in accuracy can be reduced partially  by just using distillation; see subsection~\ref{subsec:misc}.

\begin{table}[t!]
    \begin{center}
    \begin{tabular}{|l|c|c|c|c|}
    \hline
    \textbf{Model} & \textbf{\#Par. Subnets.} & \textbf{Par-Inception Top-1} & \textbf{Par-Dense. Top-1}\\
    \hline
        non-causal & 1 & 71.8 & - \\
    \hline
        sequential causal & 1 & 71.4 & 67.6 \\
    \hline
        semi-parallel causal & 5 (7) & 66.0 & 61.3 \\
    \hline
        parallel causal &10 (14)& 54.5 & 54.0 \\
    \hline
    \end{tabular}
    \end{center}
    \caption{Test accuracy as percentage for action recognition on the miniKinetics dataset~\cite{Xie2017RethinkingSF}, using networks with multi-rate clocks and temporal filters. The number of parallel subnetworks is shown in the second column. For the semi-parallel case, Par-Inception uses 5 parallel subnetworks and Par-DenseNet 7.  The non-causal, single subnetwork Par-Inception in the first row is equivalent to the I3D model \cite{carreira2017cvpr}.}
    \label{tab:sparse_results}
\end{table}

\subsection{Human keypoint localisation}
\label{sec:experiments-humanheatmap}
For this task we experimented with 5 different levels of depth-parallelism for Par-DenseNet: fully sequential and 2, 4, 7 and 14 parallel subnetworks. For Par-Inception, we used three different depth-parallelism levels: fully sequential, 5, and 10 parallel subnetworks. We employed a weighted sigmoid cross-entropy loss. Since the heatmaps contain mostly background (no-joint) pixels, we found it essential to weight the importance of the keypoint pixels in the loss -- we used a factor of 10. For evaluation, we report results on the miniKinetics test set in terms of weighted sigmoid cross-entropy loss.

\begin{figure}[h!]
\centering
\includegraphics[width=0.4\textwidth]{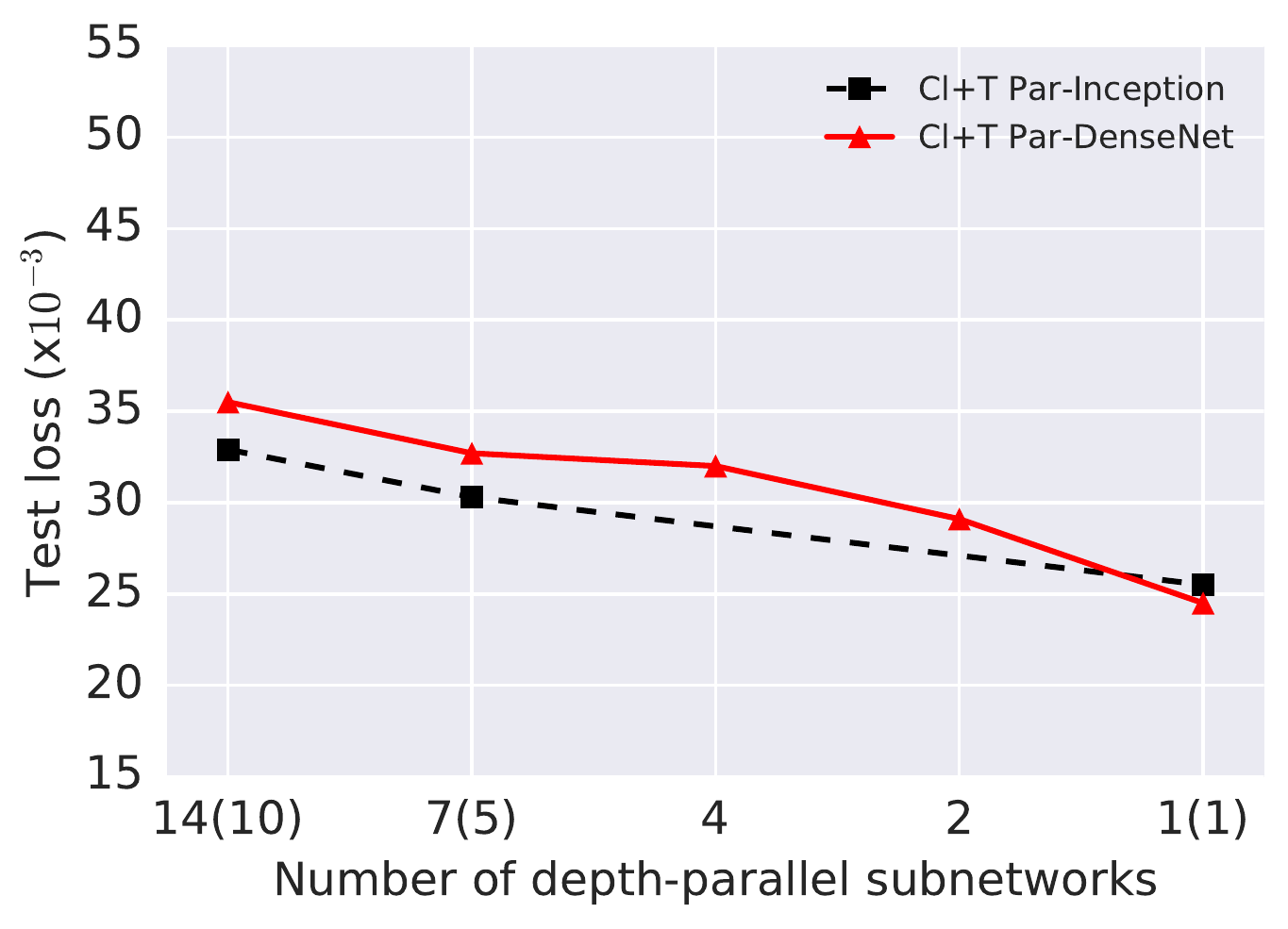}
\includegraphics[width=0.4\textwidth]{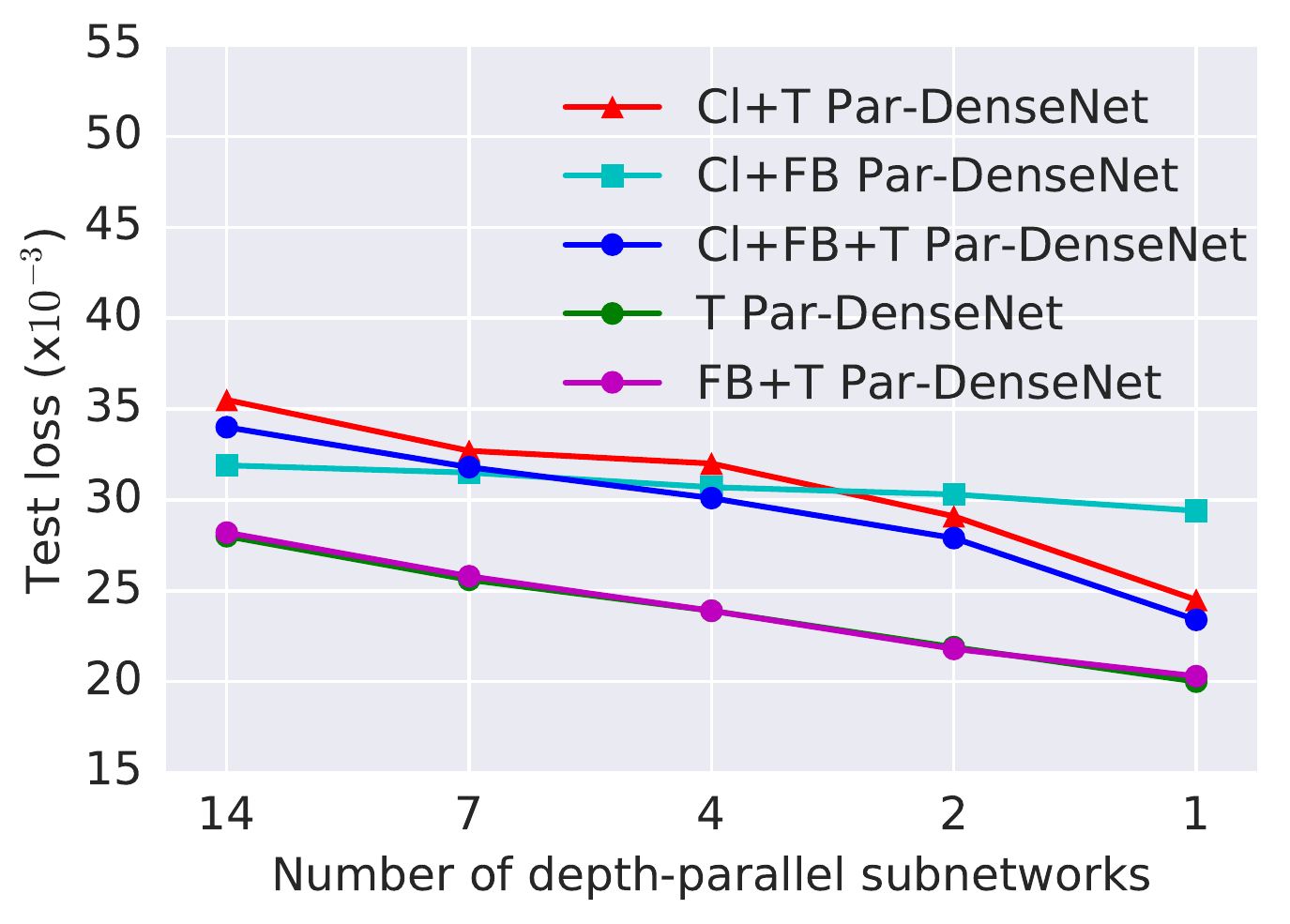}
\caption{Weighted sigmoid cross-entropy (lower is better) for human keypoint localisation on miniKinetics test set for zero  \predictionlag. ``Cl" denotes models with multi-rate clocks, ``T" -- models with temporal filters, ``FB" --  models with feedback. \textbf{Left}: Comparison between Par-Inception and Par-DenseNet for different levels of parallelism. Note that in terms of number of sequential convolutions, 14 subnetworks for Par-DenseNet are equivalent to 10 subnetworks for Par-Inception, and similar for 7(5). \textbf{Right}: Variations of Par-DenseNet. In the absence of parallelisation (1 subnetwork), the accuracy of the best models with multi-rate clocks is just slightly worse to that of a much slower sequential model. Parallelisation penalises the accuracy of models with clocks more. The basic Par-DenseNet can have up to 4 parallel subnetworks with modest drop of accuracy.}
\label{fig:kinetics_pose}
\end{figure}

Results using the pipelining connectivity with multi-rate clock models are shown in fig.~\ref{fig:kinetics_pose}, left. For both models, it can be observed that the performance improves as more layers are allowed to execute in sequence. Par-Inception has slightly better performance for higher degrees of parallelism, perhaps due to its built-in parallelism; Par-DenseNet models become better as less parallelism is used. 

\begin{figure}[h!]
\caption{Example outputs on a subset of frames one second apart from two videos of the miniKinetics test set. ``Ground truth" keypoints from the model~\cite{45946} used to automatically annotate the dataset are shown as triangles, our models predictions are shown as circles. Note that the parallel models exhibit some lag when the legs move quickly on the video on the left. Best seen zoomed on a computer screen in color. \label{fig:kinetics_pose_samples}}
\centering
{Par-DenseNet models with 14 parallel subnetworks, without clocks}\\
\includegraphics[width=0.49\textwidth]{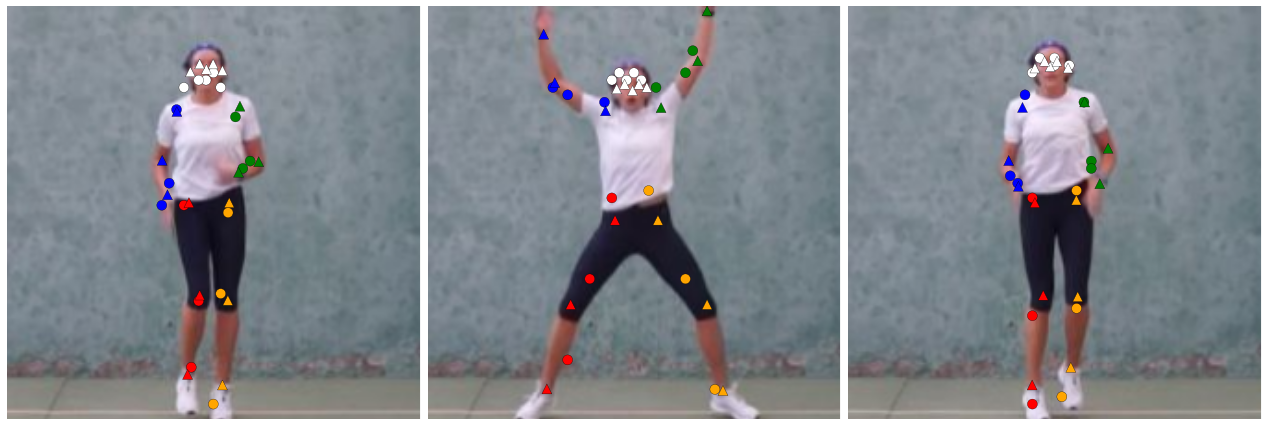}
\includegraphics[width=0.49\textwidth]{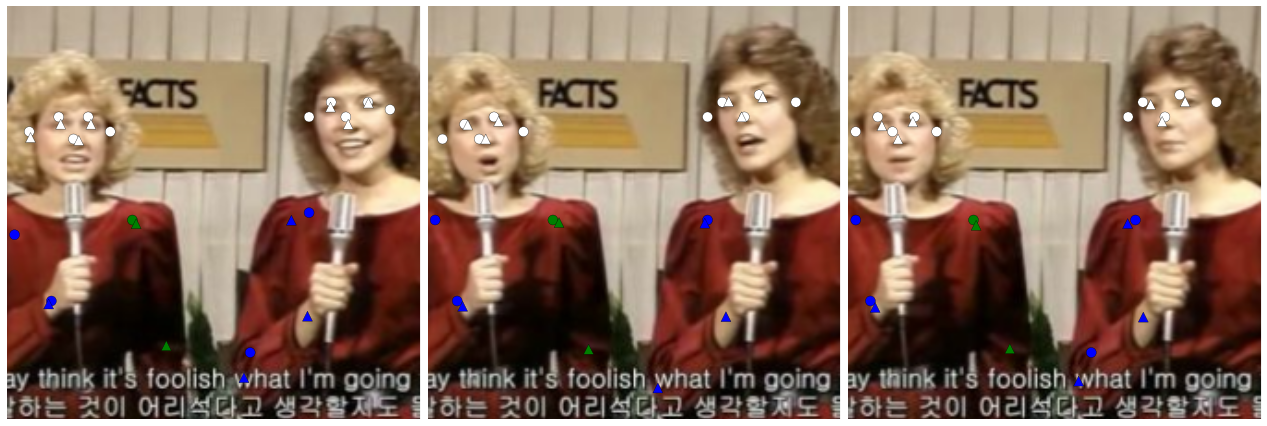}
{Fully sequential Par-DenseNet model, without clocks}\\ 
\includegraphics[width=0.49\textwidth]{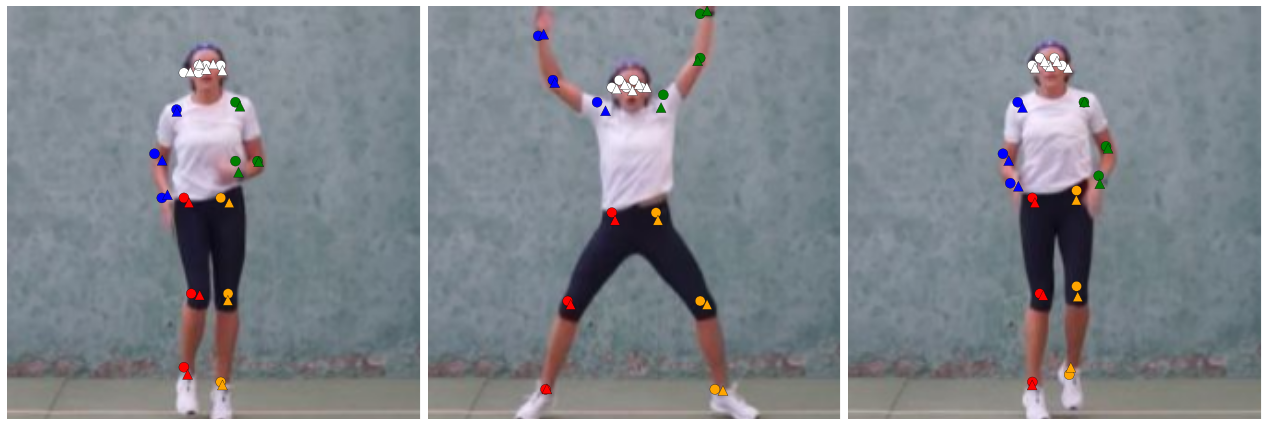}
\includegraphics[width=0.49\textwidth]{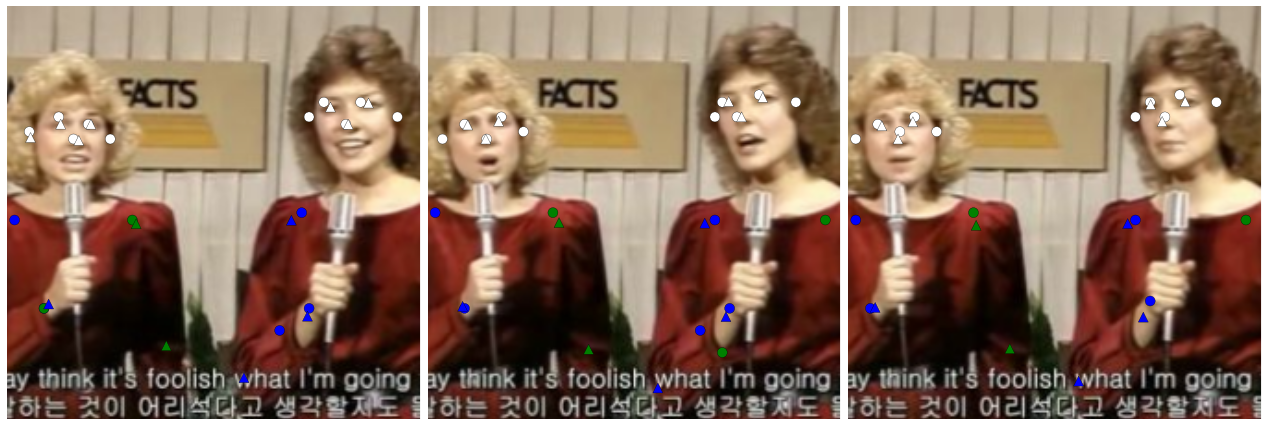}
{Par-DenseNet models with 14 parallel subnetworks, with clocks}\\
\includegraphics[width=0.49\textwidth]{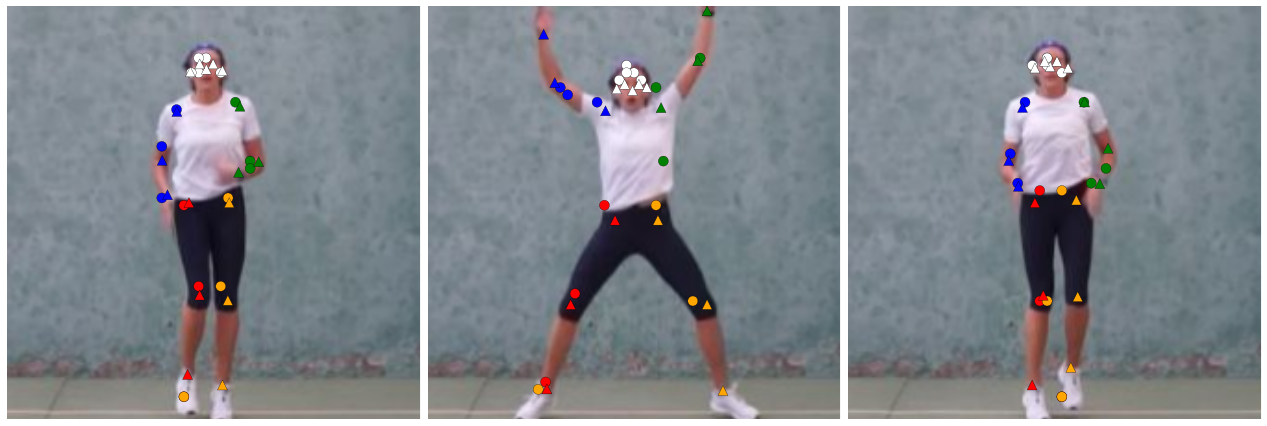}
\includegraphics[width=0.49\textwidth]{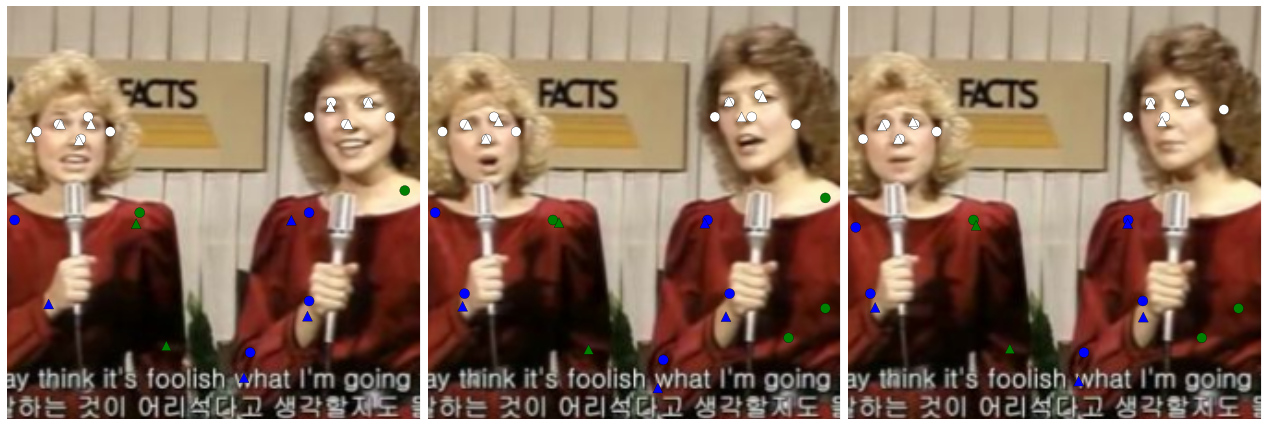}
{Fully sequential Par-DenseNet model, with clocks}\\ 
\includegraphics[width=0.49\textwidth]{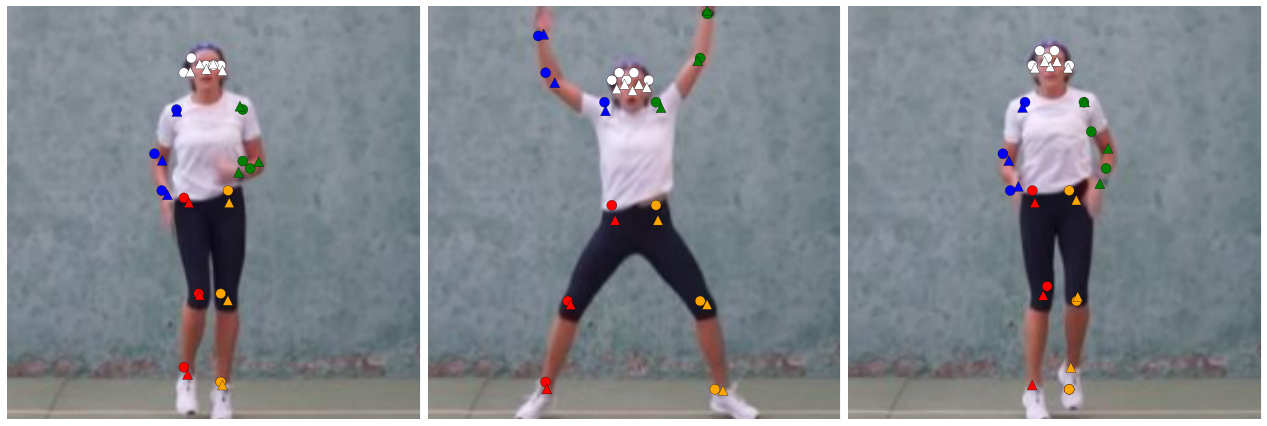}
\includegraphics[width=0.49\textwidth]{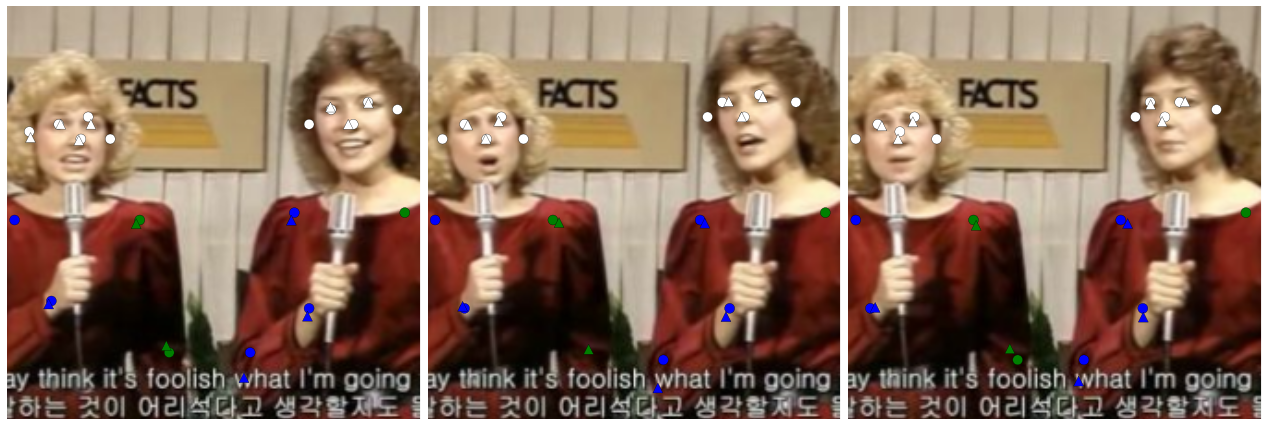}

\end{figure}

Since Par-DenseNet offers more possibilities for  parallelisation, we used it to investigate more designs, i.e.: with/without multi-rate clocks, temporal filters and feedback. The results are shown in fig.~\ref{fig:kinetics_pose}, right. Versions with temporal filters do  better than without except for the most parallel models -- these have intrinsically temporal receptive fields because of the skip connections in time, without needing explicit temporal filters. Feedback helps  slightly. Clocks degrade accuracy a little but provide big speedups (see subsection~\ref{sec:timings}). We show predictions for two test videos in fig.~\ref{fig:kinetics_pose_samples}.

\subsection{Sequential to parallel distillation}
\label{subsec:misc}
\begin{figure}[h!]
\centering
\includegraphics[width=0.4\textwidth]{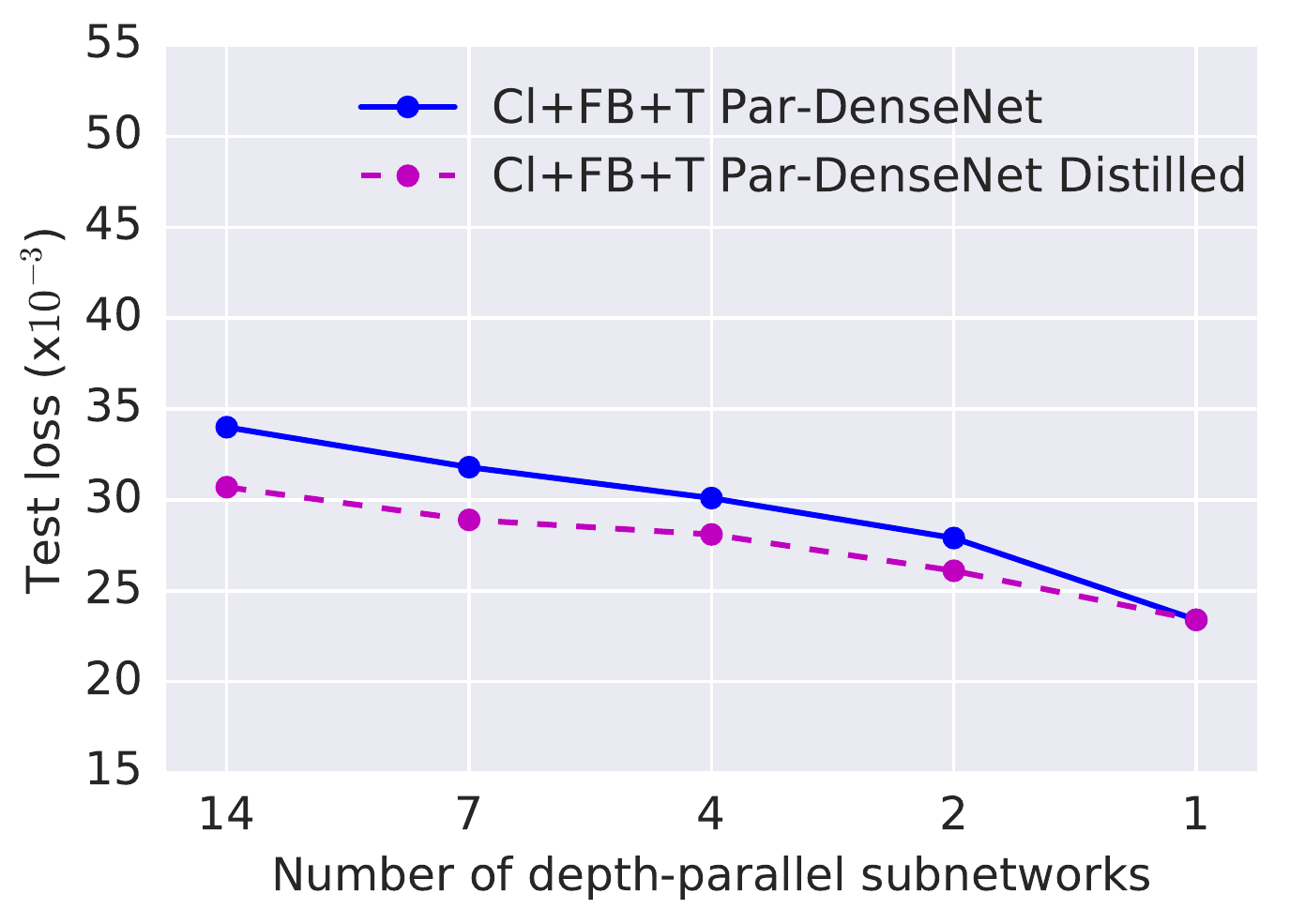}
\caption{Comparison between the weighted sigmoid cross-entropy (lower is better) of models with different levels of parallelism and the same models distilled from sequential for human keypoint localisation on miniKinetics test set for zero  \predictionlag. Results presented for a DenseNet model with multi-rate clocks (``Cl"), temporal filters (``T"), and feedback (``FB"). See text for details.}
\label{fig:distill}
\end{figure}

As mentioned in section~\ref{sec:model}, we investigated training first a sequential model, then fitting the parallel model to a subset of its activations in addition to the original loss function. This led to significant improvements for both models. The parallel causal Par-Inception model obtains a relative improvement in accuracy of about 12\%, from 54.5\% to 61.2\% for action recognition. The improvement for multi-rate Par-DenseNet model on the keypoint localisation task is shown in fig.~\ref{fig:distill}.

\subsection{Training specifically for depth-parallelism} Is it important to train a model specifically for operating in parallel mode or can we rewire a pretrained sequential model and it will work just as well at inference time? We ran an experiment where we initialiased Par-DenseNet models with different levels of parallelism with the weights from the DenseNet fully sequential model and ran inference on the miniKinetics test set. The results are shown in fig.~\ref{fig:pred-latency}, left, and indicate the importance of training with depth-parallelism enabled, so the network learns to behave predictively. We similarly evaluated the test loss of Par-DenseNet models with different levels of parallelism when initialised from a fully-parallel trained model. As expected, in this case the behaviour does not change much.

\subsection{Effect of higher \predictionlag} 
\begin{figure}[h!]
\centering
\includegraphics[width=0.4\textwidth]{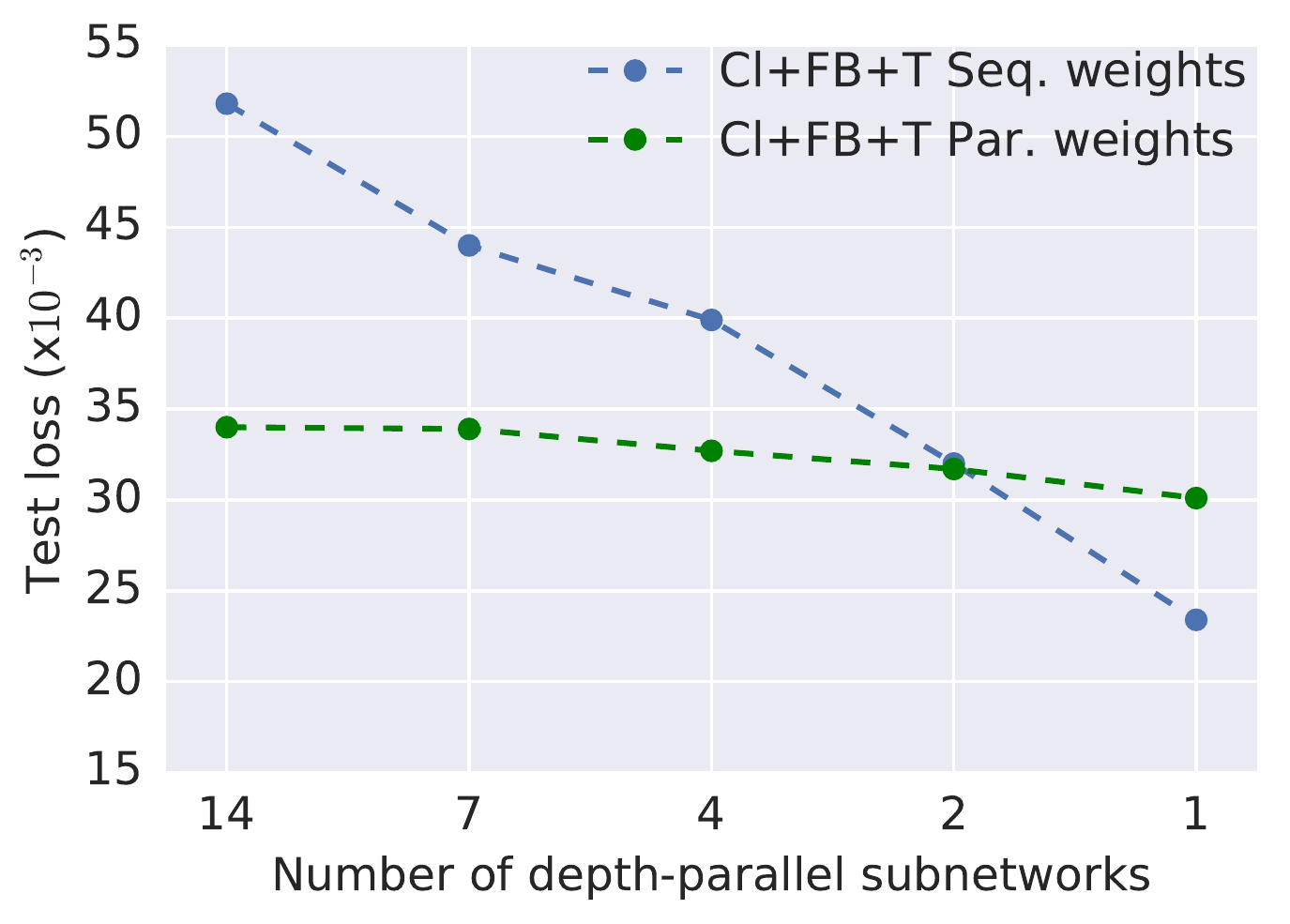}
\includegraphics[width=0.4\textwidth]{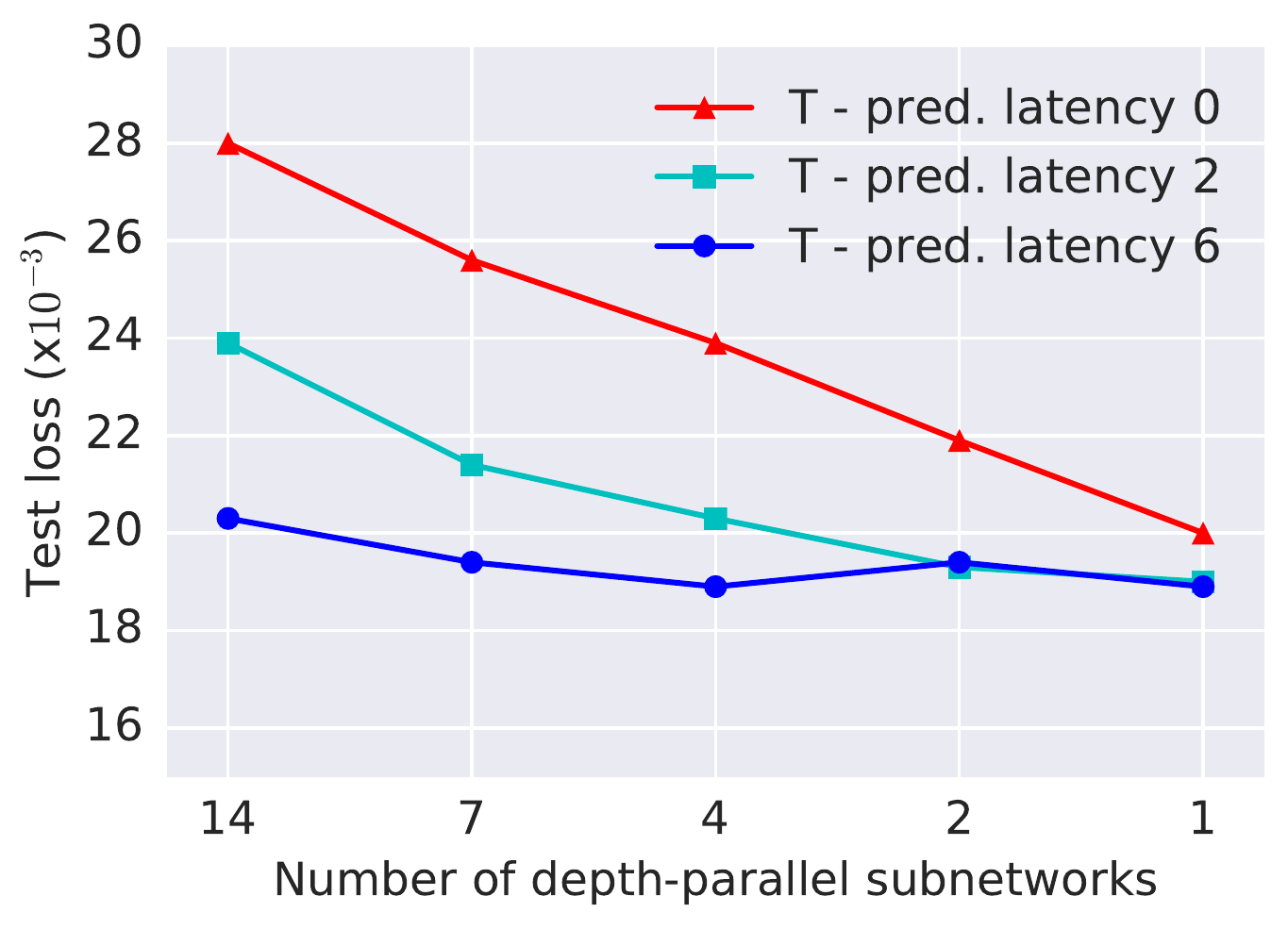}
\caption{\textbf{Left}: Seq. weights - Behaviour of Par-DenseNet with different levels of parallelism at inference time when trained with sequential connectivity. Par. weights - behaviour of Par-DenseNet with different levels of parallelism at inference time when trained with fully-parallel connectivity.  \textbf{Right}: Test loss for Par-DenseNet when prediction latency is allowed to be greater than zero.}
\label{fig:pred-latency}
\end{figure}

All the results above were obtained when training for 0 frames of \predictionlag. However, if a parallel model is several times faster than a sequential one, we can afford to introduce a \predictionlag \ greater than zero frames. Figure~\ref{fig:pred-latency}, right, shows results for Par-DenseNet models in this setting. As expected, the test loss decreases as the prediction latency increases, since more layers get to process the input frame before a prediction needs to be made. Strikingly, by using a predictive delay of 2 frames, models with up to 4 depth-parallel subnetworks are as accurate as fully sequential models with 0 frame predictive latency.

\subsection{Efficiency measurements}
\label{sec:timings}
In this section, we present the efficiency improvements achieved by the proposed models, comparing the cases with and without multi-rate clocks and with different numbers of parallel subnetworks. 
Our parallel models improve efficiency under the assumption that parallel computation resources are available. We benchmark our models on CPUs and GPUs by running inference on a CPU with 48 cores and on hosts with 2, 4, and 8 k40 GPUs, respectively. The GPUs were on the same machine to avoid network latency. For benchmarking, each model is run on 3000 frames and we average the time used to process each frame. Results are presented in table~\ref{tab:throughput}. A figure illustrating the loss in accuracy as the throughput is increased can be found in the supp. material.

Our models are implemented using TensorFlow (TF)~\cite{tensorflow2015-whitepaper}, hence: (1) when running on a multi-core CPU, we can run multiple operations in parallel and to parallelise a single operation, e.g., for conv layers. This means that the sequential model becomes faster with more cores, but only up to a certain point, when the overhead cancels out the gain from parallelism. The proposed parallel models benefit far more from having many CPU cores. (2) Multiple operations cannot run in parallel on the same GPU, hence there is little benefit in running our models on a single GPU. (3) A single operation cannot be split between GPUs. This explains why the sequential image model performance does not improve with more GPUs.

\noindent \textbf{Par-DenseNet.}
Our Par-DenseNet architecture has a total of 4+8+8+6=26 miniblocks  so when using 14 parallel subnetworks, each parallel subnetwork is made of at most 2 miniblocks. When not using multi-rate clocks, 26 miniblocks are executed for each frame resulting in 416 miniblocks executions for a sequence of 16 frames. However when using multi-rate clocks, only 86 miniblocks are executed for such a sequence, which theoretically results in a speedup of $4.8\times$. We observe some smaller speedup but this is likely to be explained by the miniblocks having different sizes. 

\noindent \textbf{Par-Inception.} Our models have 9 inception blocks. The most parallel version uses 10 parallel subnetworks: one for the initial convolutions and one for each inception block. For the sequential version, roughly a third of the time is spent on these initial convolutions. This explains why we do not observe speedups greater than 3 for the models without clocks when using more GPUs and we do not see much difference between using 4 and 8 GPU. More details together with execution timelines are included in the supp. material.
\begin{table}[t!]
    \begin{center}
    \small
    \begin{tabular}{|l|c|c|c|c|c|c|c|}
    \hline
    \textbf{Model}&\textbf{\# Par. subnets}  & \textbf{48 cores} & \textbf{2 GPUs} & \textbf{4 GPUs} & \textbf{8 GPUs}\\
    \hline
        \multicolumn{6}{|c|}{\textbf{Par-DenseNet without multi-rate clocks}} \\
\hline
sequential &1&$1.0$ & $1.0$ & $1.0$ & $1.0$ \\
\hline
semi-parallel&2 & $1.3$ & $1.6$ & $1.7$ & $1.7$ \\
\hline
semi-parallel&4& $1.8$ & $1.7$ & $2.5$ & $2.9$ \\
\hline
semi-parallel&7 & $2.2$ & $1.6$ & $2.6$ & $3.7$ \\
\hline
parallel&14& $2.6$ & $1.7$ & $2.7$ & $3.8$ \\
    \hline
        \multicolumn{6}{|c|}{\textbf{Par-DenseNet with multi-rate clocks}} \\
    \hline
sequential&1 & $2.6$ & $3.4$ & $3.4$ & $3.4$ \\
\hline
semi-parallel&2 & $3.0$ & $3.9$ & $4.0$ & $4.0$ \\
\hline
semi-parallel&4& $3.6$ & $4.5$ & $5.1$ & $5.2$ \\
\hline
semi-parallel&7 & $4.6$ & $4.5$ & $5.6$ & $6.1$ \\
\hline
parallel&14& $5.1$ & $5.0$ & $6.2$ & $7.4$ \\
    \hline
    \multicolumn{6}{|c|}{\textbf{Par-Inception without multi-rate clocks}} \\
\hline
sequential&1 & $1.0$ & $1.0$ & $1.0$ & $1.0$ \\
\hline
semi-parallel&5& $1.3$ & $1.8$ & $2.7$ & $2.7$ \\
\hline
parallel&10& $1.3$ & $1.8$ & $2.6$ & $2.6$ \\
\hline
        \multicolumn{6}{|c|}{\textbf{Par-Inception with multi-rate clocks}} \\
    \hline
sequential&1 & $2.4$ & $2.6$ & $2.6$ & $2.6$ \\
\hline
semi-parallel&5& $3.0$ & $3.4$ & $5.0$ & $5.0$ \\
\hline
parallel&10& $3.0$ & $3.4$ & $4.9$ & $5.0$ \\
    \hline
    \end{tabular} 
    \end{center}
    \caption{Throughput improvement factors for Par-DenseNet and Par-Inception models relative to a sequential network without multi-rate clocks. For Par-DenseNet the fastest model processes 7x more frames per second, whereas the fastest Par-Inception model processes 5x more frames per second; see supp. material for absolute numbers in frames per second.}
    \label{tab:throughput}
\end{table}

\section{Conclusion}

We introduced the paradigm of processing video sequences using networks that are constrained in the amount of sequential processing they can perform, with the goal of improving their efficiency. As a first exploration of this problem, we proposed a family of models where the number of sequential layers per frame is a design parameter and we evaluated how performance degrades as the allowed number of sequential layers is reduced. We have also shown that more accurate parallel models can be learned by distilling their sequential versions. We benchmarked the performance of these models considering different amounts of available parallel resources together with multi-rate clocks, and analysed the trade-off between accuracy and speedup. Interestingly, we found that the proposed design patterns can bring a speedup of up to 3 to 4x over a basic model that processes frames independently, without significant loss in performance in human action recognition and human keypoint localisation tasks. These are also general techniques -- applicable to any state-of-the-art model in order to process video more efficiently. As future work we plan to investigate further the space of possible wirings using automated strategies.

\vspace{5mm}
\noindent \textbf{Acknowledgements:} We thank Carl Doersch, Relja Arandjelovic, Evan Shelhamer, and Dominic Grewe for valuable discussions and feedback on this work.

\appendix
\chapter*{Appendix}

\section{Timing details with execution timelines}
\label{section:timeline}
\begin{figure}[h!]
\centering
\includegraphics[width=0.85\textwidth]{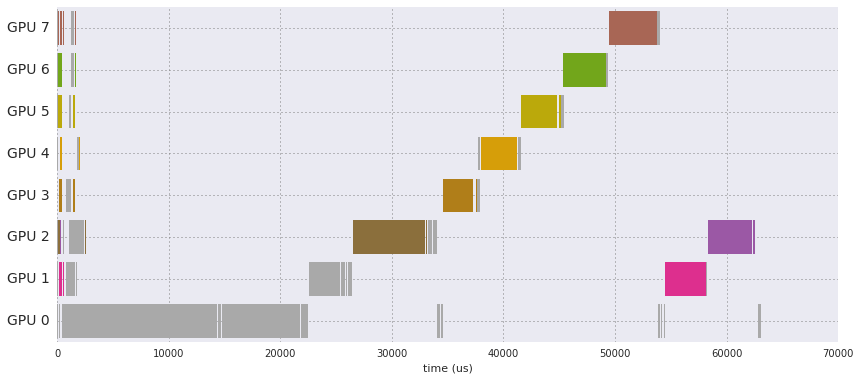}
\includegraphics[width=0.85\textwidth]{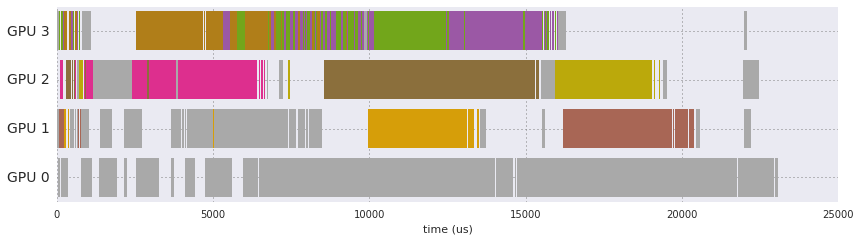}
\includegraphics[width=0.85\textwidth]{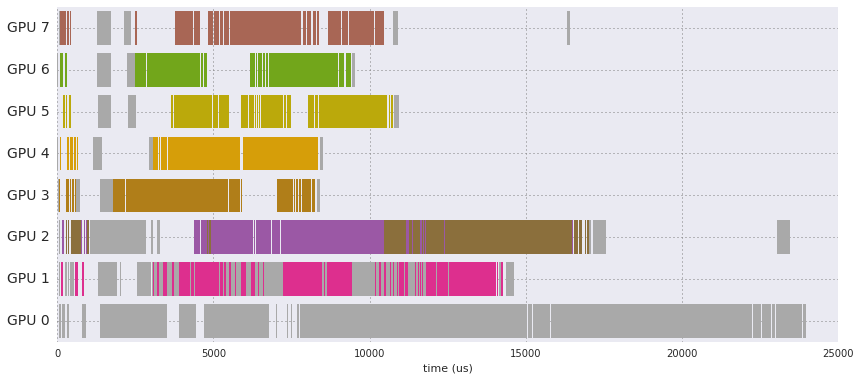}
\caption{Timeline for GPU usage for a sequential Par-Inception model on 8 GPUs at the top, a semi-parallel Par-Inception model using 4 GPUs in the middle, and a fully parallel model on 8 GPUs at the bottom. Each inception block is represented with a different color. Operations outside of inception blocks are colored in grey. Note that the timescale is different in the first picture compared to the two following ones.}
\label{fig:timeline-inception}
\end{figure}
In this section we give more details on the timing measurements.
Table~\ref{tab:sup-throughput-densenet} and table~\ref{tab:sup-throughput-inception} report the average
throughput of Par-DenseNet and Par-Inception models. GPU measurements have been done using Nvidia K-40 GPUs.
We include the throughput in frames per second in these tables. These numbers are only indicative as they
depend on the implementation (which is why they were not included in the original paper).
\begin{table}[h]
    \begin{center}
    \small
    \scalebox{0.8}{
    \begin{tabular}{|l|c|c|c|c|c|c|c|}
    \hline
    \textbf{Model}&\textbf{\# Par. Subnets}  & \textbf{48 cores} & \textbf{2 GPUs} & \textbf{4 GPUs} & \textbf{8 GPUs}\\
    \hline
        \multicolumn{6}{|c|}{\textbf{without multi-rate clocks}} \\
\hline
sequential&1 & $4.9~(1.0\times)$ & $14.1~(1.0\times)$ & $14.1~(1.0\times)$ & $14.1~(1.0\times)$ \\
\hline
semi-parallel&2 & $6.4~(1.3\times)$ & $22.3~(1.6\times)$ & $23.9~(1.7\times)$ & $23.9~(1.7\times)$ \\
\hline
semi-parallel&4 & $8.8~(1.8\times)$ & $23.5~(1.7\times)$ & $35.7~(2.5\times)$ & $40.2~(2.9\times)$ \\
\hline
semi-parallel&7 & $11.0~(2.2\times)$ & $22.3~(1.6\times)$ & $36.4~(2.6\times)$ & $51.6~(3.7\times)$ \\
\hline
parallel&14 & $12.5~(2.6\times)$ & $23.5~(1.7\times)$ & $37.7~(2.7\times)$ & $53.8~(3.8\times)$ \\
    \hline
        \multicolumn{6}{|c|}{\textbf{with multi-rate clocks}} \\
    \hline
sequential&1& $12.8~(2.6\times)$ & $48.1~(3.4\times)$ & $47.6~(3.4\times)$ & $47.8~(3.4\times)$ \\
\hline
semi-parallel&2 & $14.8~(3.0\times)$ & $55.1~(3.9\times)$ & $55.8~(4.0\times)$ & $56.2~(4.0\times)$ \\
\hline
semi-parallel&4 & $17.9~(3.6\times)$ & $63.3~(4.5\times)$ & $72.4~(5.1\times)$ & $72.8~(5.2\times)$ \\
\hline
semi-parallel&7 & $22.5~(4.6\times)$ & $64.0~(4.5\times)$ & $78.3~(5.6\times)$ & $85.4~(6.1\times)$ \\
\hline
parallel&14 & $25.2~(5.1\times)$ & $69.9~(5.0\times)$ & $87.8~(6.2\times)$ & $103.7~(7.4\times)$ \\
    \hline
    \end{tabular}}
    \end{center}
    \caption{Throughput measurements in frames per second for Par-DenseNet models.}
    \label{tab:sup-throughput-densenet}
\end{table}

\begin{table}[h]
    \begin{center}
    \scalebox{0.8}{
    \begin{tabular}{|l|c|c|c|c|c|c|c|}
    \hline
    \textbf{Model}&\textbf{\# Par. Subnets}  & \textbf{48 cores} & \textbf{2 GPUs} & \textbf{4 GPUs} & \textbf{8 GPUs}\\
    \hline
        \multicolumn{6}{|c|}{\textbf{without multi-rate clocks}} \\
\hline
sequential&1 & $6.0~(1.0\times)$ & $18.6~(1.0\times)$ & $18.0~(1.0\times)$ & $18.1~(1.0\times)$ \\
\hline
semi-parallel&5 & $7.9~(1.3\times)$ & $33.8~(1.8\times)$ & $48.7~(2.7\times)$ & $49.2~(2.7\times)$ \\
\hline
parallel&10 & $7.8~(1.3\times)$ & $33.2~(1.8\times)$ & $46.4~(2.6\times)$ & $48.1~(2.6\times)$ \\
\hline
        \multicolumn{6}{|c|}{\textbf{with multi-rate clocks}} \\
    \hline
sequential&1 & $14.3~(2.4\times)$ & $48.2~(2.6\times)$ & $47.1~(2.6\times)$ & $47.1~(2.6\times)$ \\
\hline
semi-parallel&5 & $18.1~(3.0\times)$ & $63.9~(3.4\times)$ & $90.9~(5.0\times)$ & $90.3~(5.0\times)$ \\
\hline
parallel&10 & $18.1~(3.0\times)$ & $63.7~(3.4\times)$ & $88.6~(4.9\times)$ & $90.7~(5.0\times)$ \\
    \hline
    \end{tabular}}
    \end{center}
    \caption{Throughput measurements in frames per second for Par-Inception models.}
    \label{tab:sup-throughput-inception}
\end{table}

Figure~\ref{fig:timeline-inception} represents the usage of each of the GPUs when running a sequential, a semi-parallel, and a fully-parallel Par-Inception model.
The semi-parallel and fully-parallel models run 2.6 times faster than the sequential one (the time axis has been rescaled accordingly).
Each inception block uses a different color.

We forced the sequential model to use all 8 GPUs but as expected each inception block only gets executed after the previous one in this case. It is also worth noting that the 4 branches of a given inception block are not executed in parallel, although they could be -- we tried this and did not see any noticeable speedup as one of the branches is far slower than the other three.
The inter-GPU communication overhead caused by using all 8 GPUs appears to be negligible: the frame rate we measured did not depend on the number of GPUs being used.

However when using the parallel model, all the inception blocks are able to run at the same time. The bottleneck when running with 8 GPUs is that the first three convolution layers represent roughly a third of the computation and in our model they are executed sequentially. The same bottleneck applies to the semi-parallel model, but the GPU usage
is much more balanced in this case as each of the other GPUs have at least two inception blocks to compute.
A simple workaround would be to run these three convolutional layers in parallel branches -- we plan on doing so in future work.

\begin{figure}[h!]
\centering
\includegraphics[width=0.95\textwidth]{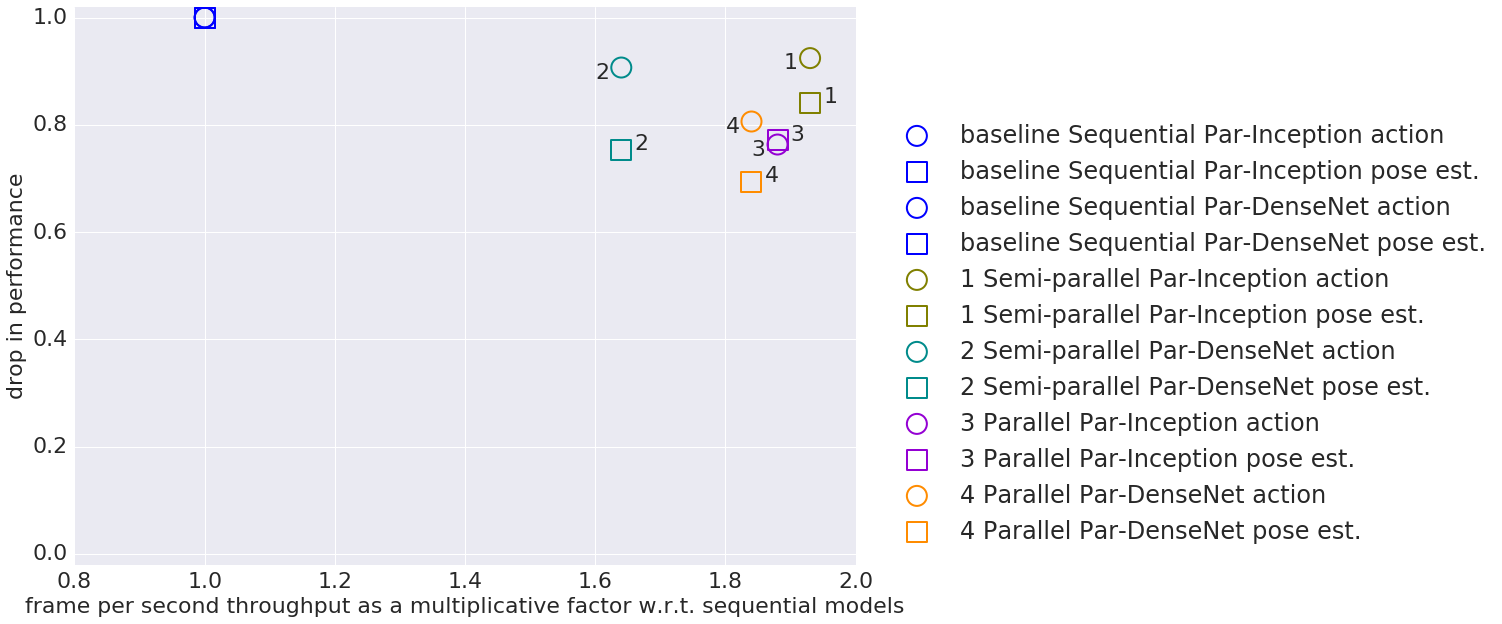}
\caption{Performance/efficiency trade-off introduced by depth-parallelising models with multi-rate clocks on 4 GPUs. Note that the baseline models use multi-rate
clocks hence the smaller speedups compared to tables~\ref{tab:sup-throughput-densenet} and~\ref{tab:sup-throughput-inception}.}
\label{fig:performance-efficiency}
\end{figure}

In order to visualise the trade-off between efficiency improvements and performance degradation, figure~\ref{fig:performance-efficiency} plots model performance with respect to efficiency for the Par-DenseNet and Par-Inception models with clocks. Round markers represent the accuracy on the action task, normalized by the accuracy obtained by the sequential model. Square markers represent the inverse of the loss on the pose estimation task, again normalized by the loss obtained by the sequential model. 

\section{Pseudocode for predictive depth-parallelism}
\label{section:pseudocode}
\begin{figure}[t!]
\begin{minipage}{0.70\textwidth}
\SetInd{0.35em}{0.25em}
\begin{algorithm}[H]
\SetKwData{Left}{left}\SetKwData{This}{this}\SetKwData{Up}{up}
\SetKwFunction{Union}{Union}\SetKwFunction{FindCompress}{FindCompress}
\SetKwInOut{Input}{input}\SetKwInOut{Output}{output}
 \Input{video frames $\{I\}$}
 \Input{number of sequential layers $k$}
 \Output{predictions $\{y\}$}
 \nosemic $n \leftarrow$ len\{layers\} \;
 clock\_rates $\leftarrow [1, 1, 1, 2, 2, 2]$ \;
 state $\leftarrow [0]$ \; \label{ln:init_state}
 $y \leftarrow$ [ ] \;
 \For{$t \leftarrow 0$ \KwTo $\emph{len}\{I\}$}{ \label{ln:unroll_loop}
  outputs $\leftarrow$ state.copy() \;  \label{ln:init_outputs}
  state.append($I[t]$) \; \label{ln:append_frame}
  \For{$d \leftarrow 0$ \KwTo $n$}{  \label{ln:traverse_net_loop}
  \If{$t \mod \emph{clock\_rates}[d] != 0$}{ \label{ln:check_tick}
  continue \; \label{ln:continue}
  }
  \eIf{$d \mod k == 0$  \label{ln:break_seq} }
  {outputs[$d$].append(layer[$d$](state[$d-1$][-2:]))\;  \label{ln:inp_state}}
  {outputs[$d$].append(layer[$d$](outputs[$d-1$][-1]))\; \label{ln:inp_output}}
 }
 state $\leftarrow$ outputs \; \label{ln:update_state}
 $y$.\text{append}(\text{classifier(outputs[-1]))} \label{ln:classif}
 }
\label{alg:model}
\end{algorithm}
\end{minipage}
\begin{minipage}{0.28\textwidth}
\includegraphics[width=\linewidth]{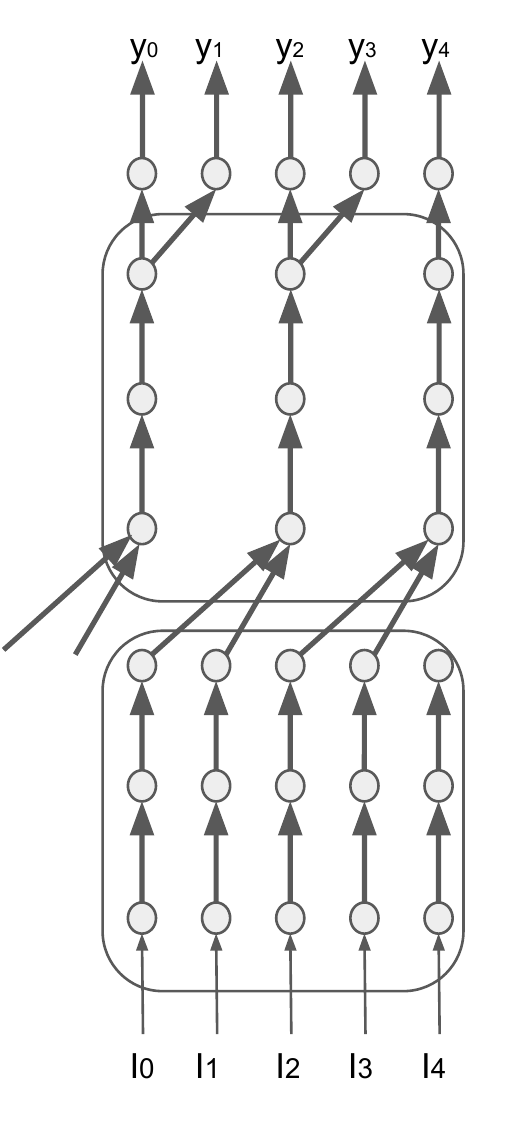}
\end{minipage}
\caption{Graph construction for predictive depth-parallel models, with multi-rate clocks.}
\end{figure}

Using the toy example in the figure below, we illustrate the construction of the TensorFlow graph for the proposed predictive depth-parallel models with multi-rate clocks in Algorithm~\ref{alg:model}. The model here has $n=6$ layers and a final classifier, and it is unrolled over $5$ time steps. The model outputs predictions $y_i$ at the same rate as the rate at which frames $I_i$ arrive. The layers are distributed into 2 parallel subnetworks, with $k=3$ sequential layers in each subnetwork, and uses a clock rate of 1 for the first subnetwork and clock rate of 2 for the second one. 

This has three implications: (1) when we break the sequence path between subnetworks, the output layer of subnetwork 1 should cache its activations for one time step, when they can be processed by the second subnetwork; (2) but because the second subnetwork ticks only every two time steps, the last layer of subnetwork 1 must actually cache its activations for two time steps, and (3) in every other time steps, the output classifier makes predictions based on stale inputs. The model is unrolled over time (line~\ref{ln:unroll_loop}), similar to an RNN, and maintains its state, more precisely maintains two steps of computation as mentioned in observation (2) above. 

At the first time step, the state is initialised to 0 (line~\ref{ln:init_state}). In every unroll step, the outputs of the network are first initialised from the state (line~\ref{ln:init_outputs}), and the current frame is appended to state (line~\ref{ln:append_frame}) to be processed by the first layer. Then the computation traverses the network in depth (line~\ref{ln:traverse_net_loop}). If the clock of a layer did not reach its tick time (line~\ref{ln:check_tick}), then the layer is simply not connected in the graph for this time step (line~\ref{ln:continue}), and its output will carry over a copy of the state to the next time step. If the clock does tick, we then need to check if the layer is to be connected in sequence (line~\ref{ln:break_seq}) -- inputs are taken from the last output of the previous layer as in standard models (line~\ref{ln:inp_output}) -- or in parallel -- inputs are taken only from the state (line~\ref{ln:inp_state}), using the two last entries in state, since the second subnetwork ticks slower. Eventually, the state is updated with the current outputs (line~\ref{ln:update_state}) and the loop is repeated for the remaining frames. The output predictions of the network are extracted from the last layer, by applying (in sequence) a classifier (line~\ref{ln:classif}). 

\section{Additional details on training setups}
\label{section:details}

We used randomly extracted subsequences of 32 frames for pose and 64 frames for action in training; the evaluation was done on the full sequences, that have up to 250 frames -- 10 seconds of video. The spatial resolution of the input frames at both training and evaluation time is $224\times 224$, obtained by random cropping at training time and central cropping for evaluation. We also randomly flipped the videos horizontally during training.

For the task of keypoint localisation, we generate dense per-frame labels by convolving the binary joint maps with a gaussian filter and obtain 17D heatmaps. Note that there can be multiple people in a single image and although this is a state-of-the-art system the labels are slightly noisy since they were automatically extracted and not verified by human experts. Also, for simplicity, we do not consider the problem of person detection, just direct keypoint localization. As a consequence of these aspects of our setup, one predicted heatmap can contain several joints of the same type belonging to different individuals. 

All our models were trained using SGD with momentum 0.9. For both tasks, the Par-Inception models were trained with initial learning rate of 0.1, and batch size of 4. For keypoint localisation, the learning rate was decreased by a factor of 10 after 35k and 55k iterations of training, whereas for action classification, it was decreased after 50k and 60k iterations. For both tasks, we ran 70k iterations of training.

The Par-DenseNet models were more memory intensive so we used a smaller batch size, 1 for keypoint localization, and 2 for classification. We trained the models with learning rate 1.0 for keypoints and 0.1 for actions, for a total of 150k iterations, lowering the learning rate by a factor of 10 at 100k iterations.

\section{Architecture details}

\label{sec:casestudies}
In this section, we explain how the proposed principles were applied on two popular image classification models: DenseNet~\cite{huang2017densely} and Inception~\cite{inception}.

\noindent \textbf{DenseNet model.} DenseNet~\cite{huang2017densely} is a  state-of-the-art model for image classification. It consists of densely-connected blocks, each composed of $b$ miniblocks. These blocks are densely connected such that every miniblock sends its activations to all the subsequent miniblocks in the same block. The model has growth-rate as a hyperparameter to control how many new features are added by each layer. Pooling layers are interleaved between the blocks. We used 4 blocks with average pooling operators in between, and growth-rate of 64. The blocks have 4, 8, 8, and 6 miniblocks (each with a 1x1 convolution followed by a 3x3 convolution). The model starts with a 7x7 convolutional layer and ends with a 3x3 heatmap prediction head for dense predictions tasks. The input to this head is a stack of skip-connections (upsampled feature maps, 56x56, from the end of each block plus the first convolutional layer). For classification tasks the head is replaced by a fully connected layer. We experimented with this model both with and without variable clock rates and temporal kernels (kernel dimension 2 along time for all layers but the first convolutional layer, where the model inputs a single image at a time). We also experimented with versions with and without feedback. When using feedback, the heatmaps predictions from the previous frame are stacked with the output of the first convolutional layer for the current frame. We trained the resulting models in all cases recurrently, restricting inputs to past frames so it behaves causally.

\noindent \textbf{Inception model.} The Inception architecture \cite{inception} is a directed graph that begins with 3 convolutional layers, followed by 9 inception blocks. Each inception block is composed of 4 parallel branches. For this model we experimented only with a version with temporal filters and variable clock rates, similar to the 3D ConvNet for action classification from the literature, I3D~\cite{carreira2017cvpr}, but transformed into a causal recurrent-style network. Masked 3D convolutions can be used for this (or, equivalently, shifting along the time dimension, as mentioned in~\cite{DBLP:journals/corr/OordDZSVGKSK16}, sec. 2.1), but we prefer the unrolling since we are interested in frame-by-frame operation. The parameters between time steps are shared, the unrolling in this case being equivalent to shifting the convolutional kernel over the time dimension when applying a 3D convolution. The variable temporal strides of I3D are incorporated by removing blocks from the unrolling graph at time steps when there is a stride gap. Similar to DenseNet model, for per-frame prediction tasks, we introduce skip-connections (upsampling the activations of each inception block and passing them through a prediction head to produce spatial heatmaps). 

\noindent \textbf{Discussion.}

 In terms of model capacity, the two models are comparable, the temporal version of Inception has 12M parameters, and the temporal version of DenseNet has 10M parameters. The length of the longest sequential path for the Inception-based model is only 22 (counted as convolutional layers), whereas for DenseNet it is 54. 
Hence there are more possible options for breaking down this path into parallel subnetworks for DenseNet than for Inception. The \informationlatency~is however  shorter for DenseNet because of its dense connectivity. The next section gives speedups for the two architectures, for different levels of parallelism.

\label{section:architecture-details}
We specify here the architectures trained for keypoint localisation or action, giving the layer structure (kernel shapes, number of channels, strides) and number of weights. ReLU and batch normalization layers are not shown, to reduce clutter, but are used as in the original image architectures.

\subsection{DenseNet: table~\ref{tab:densenet_params1}}

\begin{table}
\begin{center}
\scalebox{0.8}{
\begin{tabular}{|l|c|c|c|c|}
    \hline
     \textbf{Layer name} & \textbf{Type} & \textbf{\# channels} & \textbf{Kernel shape} & \textbf{Strides}  \\
    \hline
    \textbf{Conv3d\_1a\_7x7} & conv3d & 64 & 1x7x7 & 1, 2, 2 \\
    \hline
     \textbf{MaxPool3d\_sp} & maxpool3d & - & 1x3x3 & 1, 2, 2 \\
    \hline
    \textbf{MaxPool3d\_t} & maxpool3d & - & 2x1x1 & 2, 1, 1 \\
    \hline

    \textbf{Block\_1} & & & & \\ 
    \hspace{2mm} bottleneck\_1\_0 & conv3d & 256 & 1x1x1 & 1, 1, 1 \\
    \hspace{2mm} conv\_1\_0 & conv3d & 64 & 1x3x3 & 1, 1, 1 \\
    \hspace{2mm} bottleneck\_1\_1 & conv3d & 256 & 1x1x1 & 1, 1, 1 \\
    \hspace{2mm} conv\_1\_1 & conv3d & 64 & 1x3x3 & 1, 1, 1 \\
    \hspace{2mm} bottleneck\_1\_2 & conv3d & 256 & 1x1x1 & 1, 1, 1 \\
    \hspace{2mm} conv\_1\_2 & conv3d & 64 & 1x3x3 & 1, 1, 1 \\
    \hspace{2mm} bottleneck\_1\_3 & conv3d & 256 & 1x1x1 & 1, 1, 1 \\
    \hspace{2mm} conv\_1\_3 & conv3d & 64 & 1x3x3 & 1, 1, 1 \\
    \hspace{2mm} skip\_1 & conv3d & 16 & 1x1x1 & 1, 1, 1 \\
    \hspace{2mm} bottleneck\_1\_4 & conv3d & 168 & 1x1x1 & 1, 1, 1 \\
    \hspace{2mm} AvgPool\_1 & conv3d & - & 1x2x2 & 1, 2, 2 \\
    \hline
    \multicolumn{5}{|l|}{\hspace{2mm} output = skip\_1, AvgPool\_1}\\
    \hline

    \textbf{Block\_2} & & & & \\ 
    \hspace{2mm} bottleneck\_2\_0 & conv3d & 256 & 1x1x1 & 1, 1, 1 \\
    \hspace{2mm} conv\_2\_0 & conv3d & 64 & 1x3x3 & 1, 1, 1 \\
    \hspace{2mm} ... & ... & ... & ... & ... \\
    \hspace{2mm} conv\_2\_7 & conv3d & 64 & 1x3x3 & 1, 1, 1 \\
    \hspace{2mm} bottleneck\_2\_7 & conv3d & 256 & 1x1x1 & 1, 1, 1 \\
    \hspace{2mm} skip\_2 & conv3d & 16 & 1x1x1 & 1, 1, 1 \\
    \hspace{2mm} bottleneck\_2\_8 & conv3d & 340 & 1x1x1 & 1, 1, 1 \\
    \hspace{2mm} AvgPool\_2 & conv3d & - & 2x2x2 & 2, 2, 2 \\
    \hline
    \multicolumn{5}{|l|}{\hspace{2mm} output = skip\_2, AvgPool\_2}\\

    \hline

    \textbf{Block\_3} & & & & \\ 
    \hspace{2mm} bottleneck\_3\_0 & conv3d & 256 & 1x1x1 & 1, 1, 1 \\
    \hspace{2mm} conv\_3\_0 & conv3d & 64 & 1x3x3 & 1, 1, 1 \\
    \hspace{2mm} ... & ... & ... & ... & ... \\
    \hspace{2mm} conv\_3\_7 & conv3d & 64 & 1x3x3 & 1, 1, 1 \\
    \hspace{2mm} bottleneck\_2\_7 & conv3d & 256 & 1x1x1 & 1, 1, 1 \\
    \hspace{2mm} skip\_3 & conv3d & 16 & 1x1x1 & 1, 1, 1 \\
    \hspace{2mm} bottleneck\_3\_8 & conv3d & 426 & 1x1x1 & 1, 1, 1 \\
    \hspace{2mm} AvgPool\_3 & conv3d & - & 2x2x2 & 2, 2, 2 \\
    \hline
    \multicolumn{5}{|l|}{\hspace{2mm} output = skip\_3, AvgPool\_3}\\

    \hline

    \textbf{Block\_4} & & & & \\ 
    \hspace{2mm} bottleneck\_4\_0 & conv3d & 256 & 1x1x1 & 1, 1, 1 \\
    \hspace{2mm} conv\_4\_0 & conv3d & 64 & 1x3x3 & 1, 1, 1 \\
    \hspace{2mm} ... & ... & ... & ... & ... \\
    \hspace{2mm} conv\_4\_5 & conv3d & 64 & 1x3x3 & 1, 1, 1 \\
    \hspace{2mm} skip\_4 & conv3d & 16 & 1x1x1 & 1, 1, 1 \\
    \hline
    \multicolumn{5}{|l|}{\hspace{2mm} output = skip\_4}\\
\hline    
\multicolumn{5}{|l|}{\textbf{Upsample and concat}(MaxPool3d\_sp, skip\_1...4)}\\
\hline
\textbf{Logits} & conv3d & n\_keypoints & 1x3x3 & 1, 1, 1 \\
    \hline
    \multicolumn{5}{|l|}{\textbf{Total number of weights: 10,843,464}}\\
    \hline
\end{tabular}}
\end{center}
\caption{Parameters of Par-DenseNet models for human keypoint localization. This version has multi-rate clocks and temporal filters but no feedback -- in the version with feedback the input to "Logits" is fed back and concatenated with "MaxPool3d\_sp". The classification version does not use the "skip" layers and instead has a classification head with inputs from just block 4.}
\label{tab:densenet_params1}
\end{table}

\subsection{Inception: tables~\ref{tab:inception_params1}~-~\ref{tab:inception_params2}}

\begin{table}
\begin{center}
\scalebox{0.8}{
\begin{tabular}{|l|c|c|c|c|}
    \hline
     \textbf{Layer name} & \textbf{Type} & \textbf{\# channels} & \textbf{Kernel shape} & \textbf{Strides}  \\
    \hline
    \textbf{Conv3d\_1a\_7x7} & conv3d & 64 & 7x7x7 & 2, 2, 2 \\
    \hline
     \textbf{MaxPool3d\_2a\_3x3} & maxpool3d & - & 1x3x3 & 1, 2, 2 \\
    \hline
    \textbf{Conv3d\_2b\_1x1} & conv3d & 64 & 1x1x1 & 1, 1, 1 \\
    \hline
    \textbf{Conv3d\_2c\_3x3} & conv3d & 192 & 3x3x3 & 1, 1, 1 \\
    \hline
    \textbf{MaxPool3d\_3a\_3x3} & maxpool3d & - & 1x3x3 & 1, 2, 2 \\
    \hline
    \textbf{Mixed\_3b} & & & & \\ 
    \hspace{2mm} branch0 & conv3d & 64 & 1x1x1 & 1, 1, 1 \\
    \hspace{2mm} branch1\_0 & conv3d & 96 & 1x1x1 & 1, 1, 1 \\
    \hspace{2mm} branch1\_1 & conv3d & 128 & 3x3x3 & 1, 1, 1 \\
    \hspace{2mm} branch2\_0 & conv3d & 16 & 1x1x1 & 1, 1, 1 \\
    \hspace{2mm} branch2\_1 & conv3d & 32 & 3x3x3 & 1, 1, 1 \\
    \hspace{2mm} branch3\_0 & maxpool3d & - & 3x3x3 & 1, 1, 1 \\
    \hspace{2mm} branch3\_1 & conv3d & 32 & 1x1x1 & 1, 1, 1 \\
    \hline
    \multicolumn{5}{|l|}{\hspace{2mm} output = concat(branch0, branch1, branch2, branch3)}\\
    \hline
    \textbf{Mixed\_3c} & & & & \\ 
    \hspace{2mm} branch0 & conv3d & 128 & 1x1x1 & 1, 1, 1 \\
    \hspace{2mm} branch1\_0 & conv3d & 128 & 1x1x1 & 1, 1, 1 \\
    \hspace{2mm} branch1\_1 & conv3d & 192 & 3x3x3 & 1, 1, 1 \\
    \hspace{2mm} branch2\_0 & conv3d & 32 & 1x1x1 & 1, 1, 1 \\
    \hspace{2mm} branch2\_1 & conv3d & 96 & 3x3x3 & 1, 1, 1 \\
    \hspace{2mm} branch3\_0 & maxpool3d & - & 3x3x3 & 1, 1, 1 \\
    \hspace{2mm} branch3\_1 & conv3d & 64 & 1x1x1 & 1, 1, 1 \\
    \hline
    \multicolumn{5}{|l|}{\hspace{2mm} output = concat(branch0, branch1, branch2, branch3)}\\
    \hline
    \textbf{MaxPool3d\_4a\_3x3} & maxpool3d & - & 3x3x3 & 2, 2, 2 \\
    \hline
    \textbf{Mixed\_4b} & & & & \\ 
    \hspace{2mm} branch0 & conv3d & 192 & 1x1x1 & 1, 1, 1 \\
    \hspace{2mm} branch1\_0 & conv3d & 96 & 1x1x1 & 1, 1, 1 \\
    \hspace{2mm} branch1\_1 & conv3d & 208 & 3x3x3 & 1, 1, 1 \\
    \hspace{2mm} branch2\_0 & conv3d & 16 & 1x1x1 & 1, 1, 1 \\
    \hspace{2mm} branch2\_1 & conv3d & 48 & 3x3x3 & 1, 1, 1 \\
    \hspace{2mm} branch3\_0 & maxpool3d & - & 3x3x3 & 1, 1, 1 \\
    \hspace{2mm} branch3\_1 & conv3d & 64 & 1x1x1 & 1, 1, 1 \\
    \hline
    \multicolumn{5}{|l|}{\hspace{2mm} output = concat(branch0, branch1, branch2, branch3)}\\
    \hline
    \textbf{Mixed\_4c} & & & & \\ 
    \hspace{2mm} branch0 & conv3d & 160 & 1x1x1 & 1, 1, 1 \\
    \hspace{2mm} branch1\_0 & conv3d & 112 & 1x1x1 & 1, 1, 1 \\
    \hspace{2mm} branch1\_1 & conv3d & 224 & 3x3x3 & 1, 1, 1 \\
    \hspace{2mm} branch2\_0 & conv3d & 24 & 1x1x1 & 1, 1, 1 \\
    \hspace{2mm} branch2\_1 & conv3d & 64 & 3x3x3 & 1, 1, 1 \\
    \hspace{2mm} branch3\_0 & maxpool3d & - & 3x3x3 & 1, 1, 1 \\
    \hspace{2mm} branch3\_1 & conv3d & 64 & 1x1x1 & 1, 1, 1 \\
    \hline
    \multicolumn{5}{|l|}{\hspace{2mm} output = concat(branch0, branch1, branch2, branch3)}\\
    \hline
    \textbf{Mixed\_4d} & & & & \\ 
    \hspace{2mm} branch0 & conv3d & 128 & 1x1x1 & 1, 1, 1 \\
    \hspace{2mm} branch1\_0 & conv3d & 128 & 1x1x1 & 1, 1, 1 \\
    \hspace{2mm} branch1\_1 & conv3d & 256 & 3x3x3 & 1, 1, 1 \\
    \hspace{2mm} branch2\_0 & conv3d & 24 & 1x1x1 & 1, 1, 1 \\
    \hspace{2mm} branch2\_1 & conv3d & 64 & 3x3x3 & 1, 1, 1 \\
    \hspace{2mm} branch3\_0 & maxpool3d & - & 3x3x3 & 1, 1, 1 \\
    \hspace{2mm} branch3\_1 & conv3d & 64 & 1x1x1 & 1, 1, 1 \\
    \hline
    \multicolumn{5}{|l|}{\hspace{2mm} output = concat(branch0, branch1, branch2, branch3)}\\
    \hline
    \multicolumn{5}{|l|}{cont. on next page}\\
    \hline
\end{tabular}}
\end{center}
\caption{Parameters of Par-Inception models.}
\label{tab:inception_params1}
\end{table}

\begin{table}
\begin{center}
\scalebox{0.8}{
\begin{tabular}{|l|c|c|c|c|}
    \hline
     \textbf{Layer name} & \textbf{Type} & \textbf{\# channels} & \textbf{Kernel shape} & \textbf{Strides}  \\
    
    \hline
    \textbf{Mixed\_4e} & & & & \\ 
    \hspace{2mm} branch0 & conv3d & 128 & 1x1x1 & 1, 1, 1 \\
    \hspace{2mm} branch1\_0 & conv3d & 144 & 1x1x1 & 1, 1, 1 \\
    \hspace{2mm} branch1\_1 & conv3d & 288 & 3x3x3 & 1, 1, 1 \\
    \hspace{2mm} branch2\_0 & conv3d & 32 & 1x1x1 & 1, 1, 1 \\
    \hspace{2mm} branch2\_1 & conv3d & 64 & 3x3x3 & 1, 1, 1 \\
    \hspace{2mm} branch3\_0 & maxpool3d & - & 3x3x3 & 1, 1, 1 \\
    \hspace{2mm} branch3\_1 & conv3d & 64 & 1x1x1 & 1, 1, 1 \\
    \hline
    \multicolumn{5}{|l|}{\hspace{2mm} output = concat(branch0, branch1, branch2, branch3)}\\
    \hline
    \textbf{Mixed\_4f} & & & & \\ 
    \hspace{2mm} branch0 & conv3d & 256 & 1x1x1 & 1, 1, 1 \\
    \hspace{2mm} branch1\_0 & conv3d & 160 & 1x1x1 & 1, 1, 1 \\
    \hspace{2mm} branch1\_1 & conv3d & 320 & 3x3x3 & 1, 1, 1 \\
    \hspace{2mm} branch2\_0 & conv3d & 32 & 1x1x1 & 1, 1, 1 \\
    \hspace{2mm} branch2\_1 & conv3d & 128 & 3x3x3 & 1, 1, 1 \\
    \hspace{2mm} branch3\_0 & maxpool3d & - & 3x3x3 & 1, 1, 1 \\
    \hspace{2mm} branch3\_1 & conv3d & 128 & 1x1x1 & 1, 1, 1 \\
    \hline
    \multicolumn{5}{|l|}{\hspace{2mm} output = concat(branch0, branch1, branch2, branch3)}\\
    \hline
    \textbf{MaxPool3d\_5a\_3x3} & maxpool3d & - & 2x2x2 & 2, 2, 2 \\
    \hline
    \textbf{Mixed\_5b} & & & & \\ 
    \hspace{2mm} branch0 & conv3d & 256 & 1x1x1 & 1, 1, 1 \\
    \hspace{2mm} branch1\_0 & conv3d & 160 & 1x1x1 & 1, 1, 1 \\
    \hspace{2mm} branch1\_1 & conv3d & 320 & 3x3x3 & 1, 1, 1 \\
    \hspace{2mm} branch2\_0 & conv3d & 32 & 1x1x1 & 1, 1, 1 \\
    \hspace{2mm} branch2\_1 & conv3d & 128 & 3x3x3 & 1, 1, 1 \\
    \hspace{2mm} branch3\_0 & maxpool3d & - & 3x3x3 & 1, 1, 1 \\
    \hspace{2mm} branch3\_1 & conv3d & 128 & 1x1x1 & 1, 1, 1 \\
    \hline
    \multicolumn{5}{|l|}{\hspace{2mm} output = concat(branch0, branch1, branch2, branch3)}\\
    \hline
    \textbf{Mixed\_5c} & & & & \\ 
    \hspace{2mm} branch0 & conv3d & 384 & 1x1x1 & 1, 1, 1 \\
    \hspace{2mm} branch1\_0 & conv3d & 192 & 1x1x1 & 1, 1, 1 \\
    \hspace{2mm} branch1\_1 & conv3d & 384 & 3x3x3 & 1, 1, 1 \\
    \hspace{2mm} branch2\_0 & conv3d & 48 & 1x1x1 & 1, 1, 1 \\
    \hspace{2mm} branch2\_1 & conv3d & 128 & 3x3x3 & 1, 1, 1 \\
    \hspace{2mm} branch3\_0 & maxpool3d & - & 3x3x3 & 1, 1, 1 \\
    \hspace{2mm} branch3\_1 & conv3d & 128 & 1x1x1 & 1, 1, 1 \\
    \hline
    \multicolumn{5}{|l|}{\hspace{2mm} output = concat(branch0, branch1, branch2, branch3)}\\
    \hline
    \textbf{AvgPool3d} & avgpool3d & - & 2x7x7 & 1, 1, 1 \\
    \hline
    \textbf{Logits} & conv3d & num\_classes & 1x1x1 & 1, 1, 1 \\
    \hline
    \multicolumn{5}{|l|}{\textbf{Total number of weights: 12,501,056}}\\
    \hline
\end{tabular}}
\end{center}
\caption{(cont.) Parameters of Par-Inception models.}
\label{tab:inception_params2}
\end{table}

\bibliographystyle{splncs}
\bibliography{egbib}

\end{document}